\documentclass{article}

\PassOptionsToPackage{table}{xcolor}
\usepackage{iclr2027_conference,times}
\iclrfinalcopy
\usepackage[utf8]{inputenc}
\usepackage[T1]{fontenc}
\usepackage{url,booktabs,amsfonts,amsmath,amssymb,microtype,xcolor}
\usepackage{graphicx,algorithm,algpseudocode,amsthm,tikz}
\usepackage{siunitx}
\usepackage{flafter}
\usepackage{placeins}
\usepackage[bookmarks=false]{hyperref}
\usetikzlibrary{arrows.meta,positioning,patterns}
\definecolor{citeblue}{RGB}{0,148,0}
\definecolor{citebluee}{RGB}{0,148,148}
\definecolor{l1color}{RGB}{156,179,210}
\definecolor{l2color}{RGB}{228,170,122}
\definecolor{l4color}{RGB}{145,185,175}
\hypersetup{colorlinks=true,linkcolor=red,citecolor=citeblue,urlcolor=citebluee}

\newcommand{\method}{\textsc{ForkLeft}}
\newcommand{\KL}{D_{\mathrm{KL}}}

\definecolor{ink}{HTML}{193344}
\definecolor{forkteal}{HTML}{087F8C}
\definecolor{forkorange}{HTML}{C77726}
\definecolor{forkpurple}{HTML}{77619A}
\definecolor{forkred}{HTML}{AE5964}
\definecolor{forkblue}{HTML}{4277A5}
\hypersetup{colorlinks=true,linkcolor=ink,citecolor=ink,urlcolor=forkteal,
 pdftitle={ForkLeft: Entropy-First Rollouts for Prefix-Aligned Autoregressive-to-Diffusion Distillation},pdfauthor={Junming Liu, Jicheng Wang, Yifeng He, Hao Chen, Jianzhong Qi}}
\title{ForkLeft: Entropy-First Rollouts\\for Prefix-Aligned\\Autoregressive-to-Diffusion Distillation}
\author{Junming Liu$^{1*}$\quad Jicheng Wang$^{2*}$\quad Yifeng He$^{2}$\quad Hao Chen$^{3\dagger}$\quad Jianzhong Qi$^{1\dagger}$\\
{\normalfont $^1$University of Melbourne\quad $^2$University of California, Davis\quad $^3$University of Hong Kong}\\
{\normalfont\small\texttt{junming.liu.1@student.unimelb.edu.au}, \texttt{jianzhong.qi@unimelb.edu.au},}\\
{\normalfont\small\texttt{jicwang@ucdavis.edu}, \texttt{yfhe@ucdavis.edu}, \texttt{chenho@hku.hk}}}
\begin{document}
\maketitle
\lhead{Preprint}
{\renewcommand{\thefootnote}{}\footnotetext{$^*$Equal contribution. $^\dagger$Corresponding authors.}}
\addtocontents{toc}{\protect\setcounter{tocdepth}{-10}}
\begin{abstract}
Autoregressive Next-Token Prediction (NTP) has enabled strong reasoning capabilities in language models, while Diffusion Language Models (DLMs) offer flexible token orders and parallel generation.
We ask whether DLMs can acquire NTP-style reasoning through distillation without giving up their native generation process. Direct distillation, however, faces a fundamental mismatch: an autoregressive teacher predicts from a left prefix, whereas a DLM can condition on tokens on both sides.
We introduce \method{}, a distillation framework that resolves this mismatch by separating the student's rollout from teacher supervision. During training, the student first performs entropy-first rollouts that commit uncertain positions and expose potential forks. We then fix the resulting student prefix and distill an NTP teacher under the same context, with answer correctness determining the supervision source. At inference, the student returns to its native confidence-first parallel decoding. With Qwen3-30B-A3B-Base, \method{} improves Efficient-DLM-4B on all ten benchmarks, raising MATH500 from $72.60\%$ to $79.60\%$ and consistently outperforming three alternative designs. The gains scale with teacher strength and generalize to SDAR-4B with only $500$ updates. At matched scale, the distilled 4B and 8B students exceed the published SDAR-Chat and OPDLM models on seven benchmarks, showing that DLMs can learn NTP-style reasoning without sacrificing native parallel generation.
Code and datasets will be released upon acceptance.
\end{abstract}

\section{Introduction}
\label{sec:intro}
Autoregressive (AR) next-token prediction has produced a mature family of
strong reasoning models. Diffusion language models (DLMs) offer what AR models
lack: they resolve masked positions iteratively, write several tokens in
parallel, and choose their own generation order \citep{nie2025llada,ye2025dream}.
Distilling an AR teacher into a DLM student could transfer the reasoning
capabilities of mature AR models to this more flexible generation paradigm,
but the teacher predicts each token from a left prefix, whereas the student
predicts it from whatever it has already written on either side.

\begin{figure}[!htbp]
\centering
\resizebox{\linewidth}{!}{%
\begin{tikzpicture}[x=1cm,y=1cm,>=Latex,font=\fontsize{8}{9.3}\selectfont,
 tok/.style={draw=forkblue!50,fill=forkblue!9,rounded corners=1pt,minimum width=.49cm,minimum height=.43cm},
 target/.style={tok,draw=forkteal,fill=forkteal!10,thick},
 hidden/.style={tok,draw=black!12,text=black!22,fill=black!2},
 rightt/.style={tok,draw=forkorange,fill=forkorange!12},
 box/.style={draw=ink!30,rounded corners=2pt,minimum width=2.4cm,minimum height=.6cm},
 arr/.style={->,draw=ink!60,line width=.7pt}]
\path[use as bounding box] (0,0) rectangle (14,4.7);

\node[anchor=west,font=\bfseries] at (0,4.42) {(a) Same target, different information};
\node[anchor=west,font=\bfseries] at (7.45,4.42) {(b) Match the information used to teach};
\node[anchor=west] at (.1,3.59) {AR teacher};
\node[anchor=west] at (.1,2.48) {Diffusion student};
\foreach \y in {3.59,2.48} {
 \node[tok] at (2.8,\y) {A}; \node[tok] at (3.4,\y) {B};
 \node[target] at (4,\y) {C?};
}
\node[hidden] at (4.6,3.59) {D};\node[hidden] at (5.2,3.59) {E};
\node[rightt] at (4.6,2.48) {$\hat D$};\node[rightt] at (5.2,2.48) {$\hat E$};
\draw[->,ink!65] (3.3,3.99) to[bend left=20] (4,3.99);
\draw[->,ink!65] (3.3,2.08) to[bend right=20] (4,2.08);
\draw[->,forkorange] (4.98,2.08) to[bend left=20] (4,2.08);
\node[anchor=west,text=black!55,font=\fontsize{7.4}{8.5}\selectfont] at (2.55,3.03) {Only A and B are visible};
\node[anchor=west,text=forkorange!85!black,font=\fontsize{7.4}{8.5}\selectfont] at (2.55,1.65) {Own right-side predictions are visible too};
\node[align=left,text width=6.8cm,anchor=west] at (.1,.72)
 {A different prediction can reflect different evidence,\\not just a weaker model.};
\draw[ink!18] (7.15,.22)--(7.15,4.1);
\foreach \x/\t in {8.15/A,8.75/B} \node[tok] at (\x,3.53) {\t};
\node[anchor=west] at (9.28,3.53) {Shared student prefix};
\node[box,fill=forkorange!10] (teacher) at (9.05,2.38) {Frozen AR teacher};
\node[box,fill=forkblue!10] (student) at (12.13,2.38) {Trainable DLM};
\draw[arr] (8.75,3.24)--(8.75,2.75);
\draw[arr] (9.02,3.24)--(12.13,3.24)--(student.north);
\node[anchor=center,text=forkorange!85!black] (td) at (9.05,1.5) {$p_T(C\mid AB)$};
\node[anchor=center,text=forkblue] (sd) at (12.13,1.5) {$p_S(C\mid AB)$};
\draw[arr] (teacher.south)--(td.north);\draw[arr] (student.south)--(sd.north);
\draw[->,forkteal,thick] (10.07,1.5)--(11.08,1.5);
\node[text=forkteal,font=\bfseries] at (10.58,1.91) {Forward KL};
\node[align=left,text width=6.05cm,anchor=west] at (7.6,.55)
 {Align the teaching view. Keep bidirectional context\\for the student's native generation.};

\end{tikzpicture}}
\caption{\textbf{A common target does not imply a common prediction problem.}
(a) To predict \texttt{C}, the teacher sees \texttt{AB}; the student can also
see its own $\hat{\texttt{D}},\hat{\texttt{E}}$. Gray tokens are unavailable
to the teacher; letters are schematic. (b) We teach with the same completed
student prefix \texttt{AB}, hiding \texttt{C} and its suffix. Native
generation remains bidirectional.}
\label{fig:teaser}
\end{figure}
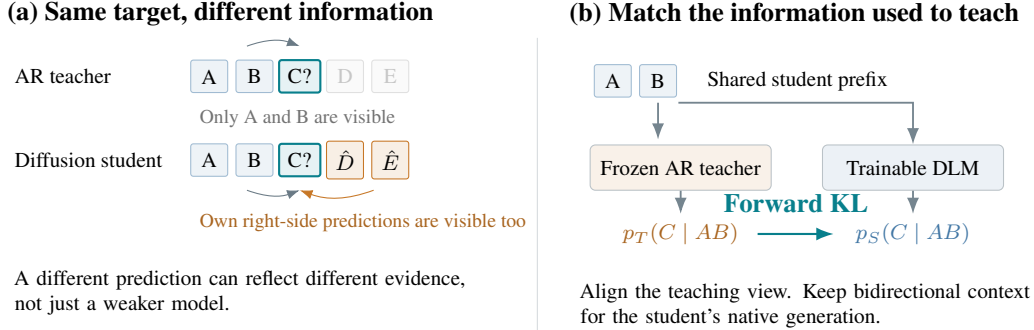

Existing DLM post-training does not address this mismatch. Random masking
trains on states that iterative decoding never visits \citep{mdpo2025}, and
confidence can become a shortcut rather than a measure of reasoning
\citep{confshortcut2026,flexibilitytrap2026}. Outcome-reward methods
such as GRPO \citep{shao2024deepseekmath} credit many parallel decisions with
one terminal scalar, so they cannot tell which token went wrong. On-policy
distillation (OPD) \citep{opdlm2026} supplies a denser signal, a teacher
distribution at every position of the student's response, but must still
decide which trajectory to teach on and under what information to compare a
causal teacher with a bidirectional student.
Borrowing both decisions from the student's inference procedure fails. The
inference sampler commits its most confident positions first, so it resolves
the hardest decision in a response, an uncertain fork, last, when its own
tokens already surround it; \citet{flexibilitytrap2026} observe this deferral
of high-entropy tokens. Trained on such trajectories, the student never faces
the fork on its own, and the teacher never sees a state at which to correct
it. Comparing the two models in the student's bidirectional view fails for a
different reason: to predict \texttt{C} in positions
\texttt{A--E}, the teacher sees \texttt{AB} while the student may also see its
own \texttt{D} and \texttt{E}, so a disagreement mixes what each model knows
with what each model sees (\autoref{fig:teaser}).

We propose \method{}, which makes both decisions for learning rather than
for decoding: \emph{expose the uncertain fork, then teach on the left}. During
training, the student commits its highest-entropy predictions first, so each
fork is decided before later context can prop it up, and the teacher never
rewrites the rollout, so an early mistake stays visible in the response. After
generation, answer correctness decides how each question is supervised, and
the teacher and the student predict each selected token from the same
student-written left prefix. These tokens are drawn uniformly along the
response rather than at the uncertain positions, so a confidently wrong token
can also be corrected. At inference, the student returns to its native
confidence-first parallel sampler.

We test the two decisions by crossing rollout order (high- or low-entropy
first) with supervision view (shared prefix or bidirectional). Of the four
configurations, entropy-first rollouts with prefix-aligned teaching
score highest on all ten benchmarks. The full method improves
Efficient-DLM-4B on every task, adds to SFT on the same questions, gains more
from stronger teachers for both student sizes, generalizes to SDAR-4B, and at
matched scale exceeds the published SDAR-Chat and OPDLM models on seven
benchmarks.

We make three contributions:
\begin{enumerate}
\item \textbf{We identify two coupled failures in AR-to-DLM transfer.}
Confidence-first rollouts hide uncertain forks, and bidirectional OPD
compares teacher and student under different information. Reversing both
yields a mean gain of $+4.93$ pp, whereas reversing only the supervision view
yields $+2.31$ pp and only the rollout order $+0.14$ pp
(\autoref{sec:ablation}).
\item \textbf{We introduce ForkLeft.} Its entropy-first student rollouts,
correctness routing, prefix-aligned teacher distributions, and confidence
ranking deliberately reverse generation priority between training and
inference without replacing student tokens with teacher tokens.
\item \textbf{We establish broad effectiveness and an empirical
teacher-scaling trend.} Across ten benchmarks, two student sizes, four
teachers, and a second student family, ForkLeft yields consistent gains; its
mean gain rises monotonically with teacher strength; and the distilled 4B and
8B students achieve state-of-the-art accuracy among published diffusion LMs of
matched scale on seven benchmarks.
\end{enumerate}

\section{Related work}
\label{sec:related}
\paragraph{Post-training diffusion language models.}
MDPO trains on progressively generated states, blockwise SFT adapts supervision
to block decoding, and reward optimization supplies outcome feedback
\citep{mdpo2025,blockwise2025}. JustGRPO constrains RL training to
left-to-right order and keeps parallel decoding at inference, avoiding a
confidence-driven flexibility trap \citep{flexibilitytrap2026}; the Confidence
Shortcut identifies a related failure \citep{confshortcut2026}, and
\citet{kim2025worst} show that masked diffusion models train under harder
orderings than the adaptive order they decode with. ForkLeft uses this training--inference asymmetry to transfer token-level
AR knowledge: uncertain decisions come first in training and last in
inference.

\paragraph{On-policy distillation.}
OPD queries a teacher on student-generated contexts. OPDLM applies it to
AR-to-DLM conversion, supervising the student's own trajectories with the
frozen AR teacher's token-level distributions \citep{opdlm2026}; diffusion
self-distillation uses later denoising states \citep{dopsd2026} or
self-generated answer suffixes \citep{luo2026dopsd}. ForkLeft designs both the
trajectory and supervision view, comparing causal teacher and bidirectional
student only on a shared student-generated prefix.

\paragraph{Learning from uncertain tokens.}
High-entropy tokens can concentrate learning signal
\citep{wang2025eighty,tip2026}. ForkLeft does not restrict loss to them:
entropy chooses when a token is committed, while uniform sampling chooses
where supervision lands. Uncertainty changes the trajectory rather than
merely reweighting its loss.
\section{Method}
\label{sec:method}

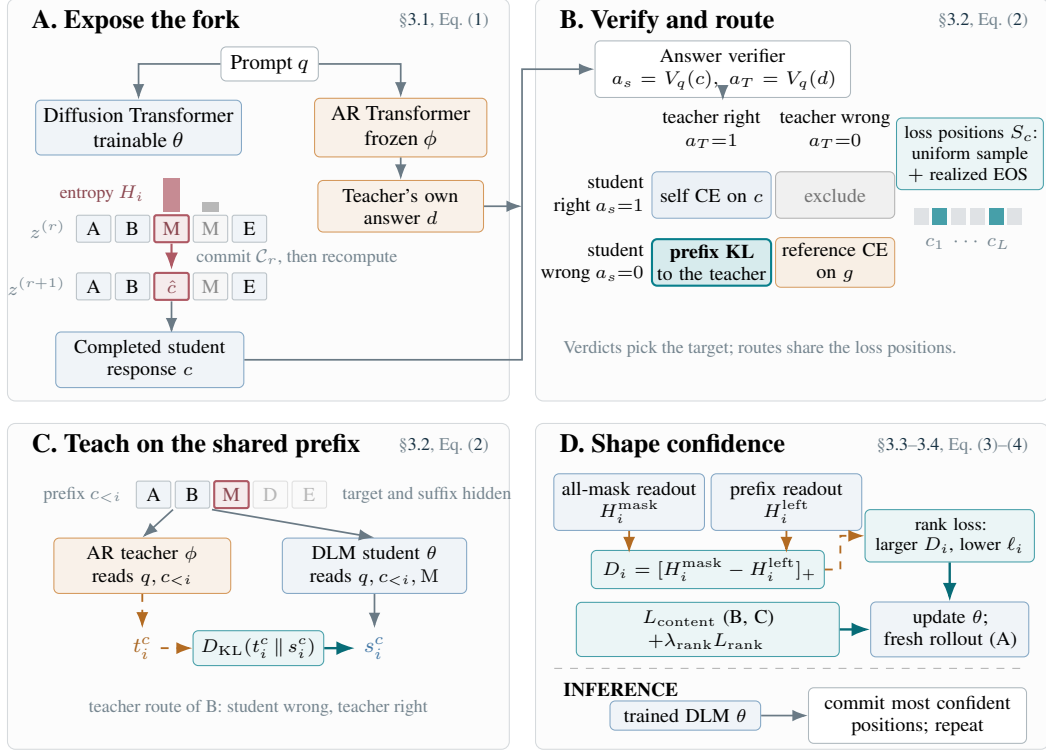
\begin{figure}[t]
\centering
\resizebox{\linewidth}{!}{%
\begin{tikzpicture}[x=1cm,y=1cm,>=Latex,
 font=\fontsize{8}{9.3}\selectfont,
 box/.style={draw=ink!35,rounded corners=2pt,fill=white,align=center,inner sep=3pt},
 student/.style={box,fill=forkblue!8,draw=forkblue!65},
 teacher/.style={box,fill=forkorange!10,draw=forkorange!75},
 loss/.style={box,fill=forkteal!8,draw=forkteal!75},
 route/.style={box,inner sep=2pt,minimum height=.62cm,text width=1.45cm,font=\fontsize{7.4}{8.5}\selectfont},
 flow/.style={->,draw=ink!65,line width=.65pt},
 fixed/.style={->,dashed,draw=forkorange!90!black,line width=.7pt},
 grad/.style={->,draw=forkteal!85!black,line width=.9pt},
 small/.style={font=\fontsize{7.4}{8.5}\selectfont},
 tiny/.style={font=\fontsize{6.8}{7.8}\selectfont},
 tok/.style={draw=ink!30,rounded corners=.8pt,fill=forkblue!8,minimum width=.46cm,minimum height=.36cm,inner sep=1pt,font=\fontsize{7.5}{8}\selectfont},
 mask/.style={tok,fill=black!3,text=black!45},
 fork/.style={tok,draw=forkred,thick,fill=forkred!12,text=forkred!80!black},
 hid/.style={tok,draw=black!12,fill=black!2,text=black!28},
 tag/.style={font=\fontsize{6.8}{7.8}\selectfont,text=ink!60,anchor=east}]
\path[use as bounding box] (-.1,-.12) rectangle (14,10.12);
\foreach \xa/\xb/\ya/\yb in {0/6.7/4.65/10.02,7.05/13.9/4.65/10.02,0/6.7/0/4.35,7.05/13.9/0/4.35}
 \draw[rounded corners=3pt,draw=ink!18,fill=black!1] (\xa,\ya) rectangle (\xb,\yb);

\node[anchor=west,font=\bfseries] at (.2,9.72) {A. Expose the fork};
\node[tag] at (6.55,9.72) {\S\ref{sec:rollout}, Eq.~\eqref{eq:commit}};
\node[box,minimum width=1.3cm] (prompt) at (3.5,9.15) {Prompt $q$};
\node[student,minimum width=2.6cm,minimum height=.6cm] (smodel) at (1.75,8.3) {Diffusion Transformer\\trainable $\theta$};
\node[teacher,minimum width=2.3cm,minimum height=.6cm] (tmodel) at (5.25,8.3) {AR Transformer\\frozen $\phi$};
\draw[flow] (prompt.west)-|(smodel.north);
\draw[flow] (prompt.east)-|(tmodel.north);
\node[tiny,anchor=east,text=ink!60] at (.9,6.95) {$z^{(r)}$};
\foreach \x/\t/\s in {1.15/A/tok,1.67/B/tok,2.19/{M}/fork,2.71/{M}/mask,3.23/E/tok} \node[\s] at (\x,6.95) {\t};
\fill[forkred!70] (2.08,7.17) rectangle (2.30,7.62);
\fill[black!28] (2.60,7.17) rectangle (2.82,7.30);
\node[tiny,anchor=east,text=forkred!80!black] at (1.95,7.42) {entropy $H_i$};
\draw[->,forkred,thick] (2.19,6.75)--(2.19,6.40);
\node[tiny,anchor=west,text=ink!60] at (2.4,6.57) {commit $\mathcal C_r$, then recompute};
\node[tiny,anchor=east,text=ink!60] at (.9,6.18) {$z^{(r+1)}$};
\foreach \x/\t/\s in {1.15/A/tok,1.67/B/tok,2.19/{$\hat c$}/fork,2.71/{M}/mask,3.23/E/tok} \node[\s] at (\x,6.18) {\t};
\draw[flow] (2.19,5.98)--(2.19,5.55);
\node[student,small,text width=2.3cm] (response) at (1.9,5.2) {Completed student\\response $c$};
\node[teacher,small,text width=2.0cm] (answerT) at (5.25,7.25) {Teacher's own\\answer $d$};
\draw[flow] (tmodel.south)--(answerT.north);

\node[anchor=west,font=\bfseries] at (7.25,9.72) {B. Verify and route};
\node[tag] at (13.75,9.72) {\S\ref{sec:fork}, Eq.~\eqref{eq:branchloss}};
\node[box,small,text width=3.2cm] (verifier) at (9.55,9.08) {Answer verifier\\$a_s=V_q(c),\ a_T=V_q(d)$};
\draw[flow] (response.east)--(6.87,5.2)--(6.87,9.08)--(verifier.west);
\draw[flow] (answerT.east)--(6.87,7.25);
\node[small,align=center] at (9.4,8.28) {teacher right\\$a_T{=}1$};
\node[small,align=center] at (11.05,8.28) {teacher wrong\\$a_T{=}0$};
\node[small,align=right,anchor=east] at (8.62,7.4) {student\\right $a_s{=}1$};
\node[small,align=right,anchor=east] at (8.62,6.5) {student\\wrong $a_s{=}0$};
\node[route,fill=forkblue!10,draw=forkblue!60] (r11) at (9.4,7.4) {self CE on $c$};
\node[route,fill=black!6,draw=black!30,text=black!55] (r10) at (11.05,7.4) {exclude};
\node[route,fill=forkteal!16,draw=forkteal,thick] (r01) at (9.4,6.5) {\textbf{prefix KL}\\to the teacher};
\node[route,fill=forkorange!12,draw=forkorange!70] (r00) at (11.05,6.5) {reference CE\\on $g$};
\draw[flow] (verifier.south)--(9.55,8.66);
\node[loss,tiny,text width=1.75cm] (sample) at (12.85,7.95) {loss positions $S_c$:\\uniform sample\\$+$ realized EOS};
\foreach \x/\f in {12.2/ink!14,12.45/forkteal!75,12.7/ink!14,12.95/ink!14,13.2/forkteal!75,13.45/ink!14} \fill[\f] (\x-.1,7.0) rectangle (\x+.1,7.22);
\node[tiny,text=ink!60] at (12.82,6.76) {$c_1\ \cdots\ c_L$};
\node[tiny,text=ink!60,anchor=west] at (7.3,5.35) {Verdicts pick the target; routes share the loss positions.};

\node[anchor=west,font=\bfseries] at (.2,4.05) {C. Teach on the shared prefix};
\node[tag] at (6.55,4.05) {\S\ref{sec:fork}, Eq.~\eqref{eq:branchloss}};
\node[tiny,anchor=east,text=ink!60] at (1.65,3.4) {prefix $c_{<i}$};
\foreach \x/\t/\s in {1.95/A/tok,2.47/B/tok,2.99/{M}/fork,3.51/D/hid,4.03/E/hid} \node[\s] at (\x,3.4) {\t};
\node[tiny,anchor=west,text=ink!60] at (4.35,3.4) {target and suffix hidden};
\node[teacher,small,text width=2.15cm] (prefixT) at (1.8,2.45) {AR teacher $\phi$\\reads $q,c_{<i}$};
\node[student,small,text width=2.25cm] (prefixS) at (4.9,2.45) {DLM student $\theta$\\reads $q,c_{<i},\mathrm M$};
\draw[flow] (2.21,3.2)--(prefixT.north);
\draw[flow] (2.73,3.2)--(prefixS.north);
\node[text=forkorange!85!black] (tt) at (1.8,1.35) {$t_i^c$};
\node[text=forkblue] (ss) at (4.9,1.35) {$s_i^c$};
\draw[fixed] (prefixT.south)--(tt.north);
\draw[flow] (prefixS.south)--(ss.north);
\node[loss,small,minimum width=1.6cm] (kl) at (3.35,1.35) {$\KL(t_i^c\,\|\,s_i^c)$};
\draw[fixed] (tt.east)--(kl.west);
\draw[grad] (kl.east)--(ss.west);
\node[tiny,text=ink!60] at (3.35,.55) {teacher route of B: student wrong, teacher right};

\node[anchor=west,font=\bfseries] at (7.25,4.05) {D. Shape confidence};
\node[tag] at (13.75,4.05) {\S\ref{sec:asym}--\ref{sec:update}, Eq.~\eqref{eq:rank}--\eqref{eq:total}};
\node[student,small,text width=1.8cm] (native) at (8.3,3.3) {all-mask readout\\$H_i^{\mathrm{mask}}$};
\node[student,small,text width=1.8cm] (reveal) at (10.4,3.3) {prefix readout\\$H_i^{\mathrm{left}}$};
\node[loss,small,text width=2.9cm] (depend) at (9.35,2.4) {$D_i=[H_i^{\mathrm{mask}}-H_i^{\mathrm{left}}]_+$};
\node[loss,tiny,text width=2.05cm] (rank) at (12.58,2.85) {rank loss:\\larger $D_i$, lower $\ell_i$};
\draw[fixed] (native.south)--(8.3,2.62);
\draw[fixed] (reveal.south)--(10.4,2.62);
\draw[fixed] (depend.east)--(11.12,2.4)--(11.12,2.85)--(rank.west);
\node[loss,small,text width=3.3cm] (total) at (9.35,1.6) {$L_{\mathrm{content}}$ (B, C) $+\lambda_{\mathrm{rank}}L_{\mathrm{rank}}$};
\node[student,small,text width=1.9cm] (update) at (12.58,1.6) {update $\theta$;\\fresh rollout (A)};
\draw[grad] (total.east)--(update.west);
\draw[grad] (rank.south)--(update.north);
\draw[dashed,black!30] (7.3,1.08)--(13.6,1.08);
\node[anchor=west,font=\bfseries\fontsize{7.4}{8.5}\selectfont] at (7.3,.8) {INFERENCE};
\node[student,small,text width=1.8cm] (testmodel) at (9.05,.45) {trained DLM $\theta$};
\node[box,small,text width=2.7cm] (decode) at (12.15,.45) {commit most confident\\positions; repeat};
\draw[flow] (testmodel.east)--(decode.west);
\end{tikzpicture}}
\caption{\textbf{ForkLeft exposes the fork, aligns the view, and keeps native inference.}
One update on the running example A--E: an entropy-first rollout, with the
frozen teacher answering only for verification (A); two verdicts select one of
four routes, with loss positions sampled uniformly (B); both models predict
the target from the same student prefix (C); prefix sensitivity ranks
confidences and decoding stays confidence-first (D). Dashed orange arrows
carry fixed targets; teal arrows carry gradients.}
\label{fig:framework}
\end{figure}

ForkLeft separates \emph{writing a response} from \emph{learning from it}.
The student writes the training response with its native bidirectional model,
committing uncertain positions first; the teacher then supervises selected
positions under the student's own left prefix; a confidence regularizer shapes
the priority the student will use after training; and inference returns to
the native confidence-first sampler. We follow one update through these four
steps in \autoref{fig:framework}: panel A is \autoref{sec:rollout}, panels B
and C are \autoref{sec:fork}, and panel D is \autoref{sec:asym} and
\autoref{sec:update}. Entropy chooses which positions the rollout commits;
uniform sampling chooses which positions receive a loss.

Consider five response positions, \texttt{A} through \texttt{E}, which denote
schematic student outputs rather than reference answers. A native state might
be \texttt{AB[M][M]E}: the student has written \texttt{A}, \texttt{B}, and
\texttt{E}, and may use \texttt{E} to generate \texttt{C}. Once the response
is complete, we mask \texttt{C} and its suffix and condition the teacher only
on the completed prefix \texttt{AB} to predict \texttt{C}. This gives the
teacher the same target under a well-defined left-to-right context while
leaving the student's native bidirectional generation unchanged.

\subsection{Expose the fork: entropy-first student rollouts}
\label{sec:rollout}

We show one rollout pass in \autoref{fig:framework}A. Let $P$ be a tokenized
question, $\theta$ the student parameters, and
$\phi$ the frozen teacher parameters. In a generation segment, the student
starts with masked response positions. Before pass $r$, let $z^{(r)}$
be the current state, $\mathcal M_r$ its unresolved positions, and $R$
the pass budget. One native forward pass produces a token distribution
$p^{\mathrm{den}}_{\theta,i}(\cdot\mid P,z^{(r)})$ at every unresolved
position. We prioritize distributions with greater entropy $H$:
\begin{equation}
\mathcal C_r=\operatorname{TopK}_{i\in\mathcal M_r}
 \!\Big(\underbrace{H\!\big(p^{\mathrm{den}}_{\theta,i}
 (\cdot\mid P,z^{(r)})\big)}_{\text{uncertainty at position } i};\;
 \underbrace{k_r}_{\text{commits this pass}}\Big),
\qquad
k_r=\left\lceil\frac{|\mathcal M_r|}{R-r+1}\right\rceil.
\label{eq:commit}
\end{equation}
Here $\mathcal C_r$ contains the $k_r$ positions to commit. Each receives
its argmax token from that same forward pass. The next pass recomputes
predictions after all these commitments become visible. Crucially, the
teacher neither chooses nor replaces these tokens. If an uncertain student
prediction creates a wrong branch, later student tokens are generated on
that branch; the completed response therefore records the consequence that
on-policy supervision must repair. This order is the first axis of the
comparison in \autoref{sec:ablation}: with the same prefix-aligned loss,
committing high-entropy positions first adds $2.62$ pp on average over
committing low-entropy positions first.

In our example, C commits before D when it is more uncertain; if both commit
in one pass, neither informs the other within that pass. We write the
completed response as $c=(c_1,\ldots,c_L)$.

\subsection{Verify, route, and teach on the shared prefix}
\label{sec:fork}

We route each completed response as in \autoref{fig:framework}B and teach
in the view of \autoref{fig:framework}C. To teach C, both models receive AB and neither receives C, D,
or E. A and B come from the completed response and need not both have been
visible when C was committed, so prefix supervision does not impose
left-to-right generation.

For any continuation $y$, write the teacher distribution as
$t_i^y=p_\phi(\cdot\mid P,y_{<i})$ and the student's prefix readout as
$s_i^y=p^\ell_{\theta,i}(\cdot\mid P,y_{<i})$; we compare normalized
distributions over the same token events (the output-head adapter is in
the appendix). We form $S_y$ from
uniformly sampled eligible positions plus the realized end-of-sequence (EOS)
index, if present. Entropy does not select loss positions, so a confident
error can still receive supervision.

A final-answer verifier marks the student response and a separate greedy
teacher response $d$ as correct or not, $a_s,a_T\in\{0,1\}$, and the pair of
verdicts selects one of four routes (\autoref{fig:framework}B). When both
answers are correct, the student already has a working solution, so we
reinforce its own tokens instead of pulling them toward the teacher. When only
the teacher is correct, we match the teacher's prefix distributions; this is
the one route that uses the teacher (\autoref{sec:subset} isolates it). When both are wrong, neither model offers
a reliable target, so we supervise an available reference $g$ under its own
prefixes. When only the student is correct, we drop the question, because
imitating a wrong teacher would unlearn a correct answer. Both asymmetric
routes are common: on GSM8K the evaluated response pairs fall on them $206$ and
$105$ times (\autoref{app:diagnostics}). On the teacher route we query the
teacher on the student's AB prefix, which may differ from the prefix in the
teacher's own answer. These choices give
\begin{equation}
L_{\mathrm{content}}(q)=
\begin{cases}
 \operatorname{mean}_{i\in S_c}[-\log s_i^c(c_i)],
   &(a_s,a_T)=(1,1)\text{: both correct},\\
 \operatorname{mean}_{i\in S_c}[\KL(t_i^c\|s_i^c)],
   &(a_s,a_T)=(0,1)\text{: teacher only},\\
 \operatorname{mean}_{i\in S_g}[-\log s_i^g(g_i)],
   &(a_s,a_T)=(0,0)\text{: both wrong, } g\text{ available},\\
 0,&\text{otherwise}.
\end{cases}
\label{eq:branchloss}
\end{equation}
Each mean divides by the selected-set size; an empty set contributes zero.
Reference targets use reference prefixes. Exclusion removes both losses.
The prefix view in the second route is the second axis of
\autoref{sec:ablation}: under the same entropy-first rollouts, replacing the
student's bidirectional readout by the prefix readout raises the mean gain
from $0.14$ to $4.93$ pp.

\subsection{Shape confidence with a prefix-sensitivity signal}
\label{sec:asym}

We add a second, training-only objective (\autoref{fig:framework}D) that
regularizes how confidently the student predicts selected positions.
For each $i\in S_c$, compare an all-mask readout with the completed-prefix
readout. Their positive entropy contrast is
$D_i=\operatorname{stopgrad}([H_i^{\mathrm{mask}}-H_i^{\mathrm{left}}]_+)$.
This quantity measures sensitivity to the two input constructions. It is
a training proxy, rather than a token-correctness label.

Let $\ell_i=\max_u\log s_i^c(u)$ be the live prefix confidence and
$\mathcal P_q=\{(i,j):i,j\in S_c,\ D_i>D_j+\mu\}$ the qualifying pairs.
With positive margins $\mu$ and $\delta$, we use
\begin{equation}
L_{\mathrm{rank}}(q)=\frac{1}{|\mathcal P_q|}
 \sum_{(i,j)\in\mathcal P_q}
 \underbrace{[\ell_i-\ell_j+\delta]_+}_{\text{penalize } i \text{ more confident than } j},
\qquad L_{\mathrm{rank}}(q)=0\ \text{if }\mathcal P_q=\varnothing.
\label{eq:rank}
\end{equation}
For example, if $D_C>D_D+\mu$, this loss favors
$\ell_C\leq\ell_D-\delta$: a larger entropy contrast receives lower
prefix confidence. The scores
and pairs are detached; gradients pass through $\ell_i$. Ranking uses the
student response on every retained route, including reference supervision.
The intent runs in three steps: the term is a training loss only; it lowers
prefix confidence where revealing the prefix changes the prediction most; and
because inference commits by confidence, the trained student should then
defer those positions until more context is available. The last step is
learned through shared parameters rather than enforced, which is why we call
the term a surrogate.

\subsection{Training objective and native inference}
\label{sec:update}

For a question batch $\mathcal B$, let $\mathcal B^+$ exclude the
student-correct, teacher-wrong cases. We optimize
\begin{equation}
L(\theta)=\frac{1}{|\mathcal B|}\sum_{q\in\mathcal B^+}
 \left[L_{\mathrm{content}}(q)
       +\lambda_{\mathrm{rank}}L_{\mathrm{rank}}(q)\right].
\label{eq:total}
\end{equation}
The original batch size remains the denominator, so excluded questions
contribute zero, and $\lambda_{\mathrm{rank}}$ balances the two terms.
Responses, verdicts, selected positions, and teacher targets stay fixed
during an update, and each update collects fresh responses
(\autoref{alg:method}).

At inference, the student ranks unresolved positions by maximum token
probability, commits the selected batch, and recomputes on its native
bidirectional state; this confidence-first rule differs from minimum-entropy
ranking, and only the student runs. The reversed order is deliberate:
entropy-first training exposes an uncertain fork before its continuation can
conceal it, whereas confidence-first inference lets the same position use as
much reliable context as the sampler can provide.

\section{Experiments}
\label{sec:results}
We ask four questions in turn. Does the complete method improve a student
that keeps its native diffusion inference (\autoref{sec:mainresults})? Do
entropy-first rollouts and prefix-aligned teaching each matter, and do they
work best together (\autoref{sec:ablation})? Do teacher distributions add
information beyond SFT on the same questions (\autoref{sec:subset})? And do
the gains grow with stronger teachers (\autoref{sec:scale}) and generalize to a
second student family (\autoref{sec:generalize}), and how do the distilled
students compare with published diffusion models of the same size
(\autoref{sec:external})?

\subsection{Evaluation setup}
\label{sec:setup}
We trained Efficient-DLM-4B \citep{efficientdlm2025} on math-domain
questions for $\num{2000}$ updates with Qwen3-30B-A3B-Base
\citep{qwen32025}, using a $\num{1024}$-token response budget, $32$-token
blocks, and at most $32$ passes per block. Our primary baseline is the same
student checkpoint before distillation, which we call the \emph{base}. We
also evaluate EDLM-8B, distill SDAR-4B with the same teacher, and compare
with the published SDAR-Chat and OPDLM models at 4B and 8B. Qwen3-30B-A3B has 30B total and about 3B active parameters
per token; we abbreviate it to Qwen3-30B.

We evaluate ten benchmarks with two protocols. GSM8K, MATH500, AIME24,
AIME25, LiveMathBench-Hard (LMB-H), and MT-AIME2024 use generated answers
from each diffusion family's native sampler with a $\num{2048}$-token budget
($\num{4096}$ for MT-AIME). GPQA-Diamond, MLogiQA, CEval, and MMLU use
masked-answer-slot likelihoods over the offered answer letters. Both
protocols use the OPDLM prompt template and answer scorer. We compute
accuracy and changes from correct counts before rounding, in percentage
points (pp; details in \autoref{app:implementation}).

\subsection{ForkLeft improves the student across all ten benchmarks}
\label{sec:mainresults}
The first test is whether teaching under shared prefixes improves a student
that keeps its native evaluation procedure. It does: EDLM-4B improves on all
ten benchmarks (\autoref{tab:main}). MATH500 rises from $72.60\%$ to
$79.60\%$, GSM8K from $82.71\%$ to $85.22\%$, and MT-AIME from $8.55\%$ to
$15.82\%$; the student need not use a causal sampler at test time.
\begin{table}[!htbp]
\centering
\caption{\textbf{ForkLeft improves EDLM-4B on all ten benchmarks.} The same student before and after $\num{2000}$ updates with Qwen3-30B-A3B-Base; accuracies in percent, gains from unrounded counts; AR reference models in \autoref{tab:referencecounts}.}
\label{tab:main}
{\footnotesize\setlength{\tabcolsep}{3.2pt}
\begin{tabular}{lrrrr}
\toprule
Benchmark & Items & Base & ForkLeft & Gain (pp)\\
\midrule
GSM8K & \num{1319} & 82.71 & \cellcolor{forkteal!9}\textbf{85.22} & +2.50\\
MATH500 & \num{500} & 72.60 & \cellcolor{forkteal!9}\textbf{79.60} & +7.00\\
AIME24 & \num{30} & 10.00 & \cellcolor{forkteal!9}\textbf{16.67} & +6.67\\
AIME25 & \num{30} & 6.67 & \cellcolor{forkteal!9}\textbf{13.33} & +6.67\\
LMB-H & \num{45} & 8.89 & \cellcolor{forkteal!9}\textbf{11.11} & +2.22\\
\bottomrule
\end{tabular}\hspace{1.0em}
\begin{tabular}{lrrrr}
\toprule
Benchmark & Items & Base & ForkLeft & Gain (pp)\\
\midrule
GPQA-D & \num{198} & 30.81 & \cellcolor{forkteal!9}\textbf{32.32} & +1.52\\
MLogiQA & \num{800} & 45.63 & \cellcolor{forkteal!9}\textbf{53.13} & +7.50\\
CEval & \num{1346} & 70.73 & \cellcolor{forkteal!9}\textbf{73.70} & +2.97\\
MMLU & \num{14042} & 71.04 & \cellcolor{forkteal!9}\textbf{76.05} & +5.01\\
MT-AIME & \num{1650} & 8.55 & \cellcolor{forkteal!9}\textbf{15.82} & +7.27\\
\bottomrule
\end{tabular}}
\end{table}

The gains also hold under multiple-choice likelihood scoring, which measures
answer preference rather than generation: MLogiQA rises from $45.63\%$ to
$53.13\%$ and MMLU from $71.04\%$ to $76.05\%$.

\begin{figure}[!htb]
\centering
\resizebox{\linewidth}{!}{%
\begin{tikzpicture}[x=1cm,y=1cm,font=\fontsize{8}{9}\selectfont]
\path[use as bounding box] (-.1,-.85) rectangle (15.6,4.28);
\node[anchor=west,font=\bfseries] at (0.55,4.05) {(a) Generated answers};
\draw[black!12] (0.55,0.2)--(5.2,0.2);\node[anchor=east] at (0.43000000000000005,0.2) {0};
\draw[black!12] (0.55,1.01)--(5.2,1.01);\node[anchor=east] at (0.43000000000000005,1.01) {2};
\draw[black!12] (0.55,1.82)--(5.2,1.82);\node[anchor=east] at (0.43000000000000005,1.82) {4};
\draw[black!12] (0.55,2.6300000000000003)--(5.2,2.6300000000000003);\node[anchor=east] at (0.43000000000000005,2.6300000000000003) {6};
\draw[black!12] (0.55,3.4400000000000004)--(5.2,3.4400000000000004);\node[anchor=east] at (0.43000000000000005,3.4400000000000004) {8};
\draw[black!50] (0.55,.2)--(5.2,.2);
\node[anchor=north,font=\fontsize{7}{8}\selectfont] at (0.55,.08) {0};
\node[anchor=north,font=\fontsize{7}{8}\selectfont] at (1.7125000000000001,.08) {500};
\node[anchor=north,font=\fontsize{7}{8}\selectfont] at (2.875,.08) {1000};
\node[anchor=north,font=\fontsize{7}{8}\selectfont] at (5.2,.08) {2000};
\node at (2.875,-.59) {Optimizer updates};
\node[rotate=90] at (-0.010000000000000009,1.85) {$\Delta$ accuracy (pp)};
\draw[forkblue,line width=.9pt] (0.5500,0.2000)--(0.7825,0.3842)--(1.1313,0.6299)--(1.7125,0.8755)--(2.8750,1.0904)--(5.2000,1.2133);
\fill[forkblue] (0.7825,0.3842) circle(1.65pt);
\fill[forkblue] (1.1313,0.6299) circle(1.65pt);
\fill[forkblue] (1.7125,0.8755) circle(1.65pt);
\fill[forkblue] (2.8750,1.0904) circle(1.65pt);
\fill[forkblue] (5.2000,1.2133) circle(1.65pt);
\draw[forkblue!65,thin] (5.24,1.2133)--(5.5200000000000005,1.2500);
\node[anchor=west,text=forkblue,font=\fontsize{7.5}{8.5}\selectfont] at (5.55,1.2500) {GSM8K};
\draw[forkteal,line width=.9pt] (0.5500,0.2000)--(0.7825,0.6050)--(1.1313,1.1720)--(1.7125,1.9010)--(2.8750,2.5490)--(5.2000,3.0350);
\fill[forkteal] (0.7825,0.6050) circle(1.65pt);
\fill[forkteal] (1.1313,1.1720) circle(1.65pt);
\fill[forkteal] (1.7125,1.9010) circle(1.65pt);
\fill[forkteal] (2.8750,2.5490) circle(1.65pt);
\fill[forkteal] (5.2000,3.0350) circle(1.65pt);
\draw[forkteal!65,thin] (5.24,3.0350)--(5.5200000000000005,3.2200);
\node[anchor=west,text=forkteal,font=\fontsize{7.5}{8.5}\selectfont] at (5.55,3.2200) {MATH500};
\draw[forkorange,line width=.9pt] (0.5500,0.2000)--(0.7825,0.2000)--(1.1313,1.5500)--(1.7125,1.5500)--(2.8750,1.5500)--(5.2000,2.9000);
\fill[forkorange] (0.7825,0.2000) circle(1.65pt);
\fill[forkorange] (1.1313,1.5500) circle(1.65pt);
\fill[forkorange] (1.7125,1.5500) circle(1.65pt);
\fill[forkorange] (2.8750,1.5500) circle(1.65pt);
\fill[forkorange] (5.2000,2.9000) circle(1.65pt);
\draw[forkorange!65,thin] (5.24,2.9000)--(5.5200000000000005,2.7800);
\node[anchor=west,text=forkorange,font=\fontsize{7.5}{8.5}\selectfont] at (5.55,2.7800) {AIME24};
\draw[forkpurple,line width=.9pt,dashed] (0.5500,0.2000)--(0.7825,0.2000)--(1.1313,1.5500)--(1.7125,1.5500)--(2.8750,2.9000)--(5.2000,2.9000);
\fill[forkpurple] (0.7825,0.2000) circle(1.65pt);
\fill[forkpurple] (1.1313,1.5500) circle(1.65pt);
\fill[forkpurple] (1.7125,1.5500) circle(1.65pt);
\fill[forkpurple] (2.8750,2.9000) circle(1.65pt);
\fill[forkpurple] (5.2000,2.9000) circle(1.65pt);
\draw[forkpurple!65,thin] (5.24,2.9000)--(5.5200000000000005,2.4600);
\node[anchor=west,text=forkpurple,font=\fontsize{7.5}{8.5}\selectfont] at (5.55,2.4600) {AIME25};
\draw[forkred,line width=.9pt] (0.5500,0.2000)--(0.7825,0.2000)--(1.1313,0.2000)--(1.7125,1.1000)--(2.8750,1.1000)--(5.2000,1.1000);
\fill[forkred] (0.7825,0.2000) circle(1.65pt);
\fill[forkred] (1.1313,0.2000) circle(1.65pt);
\fill[forkred] (1.7125,1.1000) circle(1.65pt);
\fill[forkred] (2.8750,1.1000) circle(1.65pt);
\fill[forkred] (5.2000,1.1000) circle(1.65pt);
\draw[forkred!65,thin] (5.24,1.1000)--(5.5200000000000005,0.8100);
\node[anchor=west,text=forkred,font=\fontsize{7.5}{8.5}\selectfont] at (5.55,0.8100) {LMB-H};
\draw[ink,line width=.9pt,dashed] (0.5500,0.2000)--(0.7825,0.5191)--(1.1313,1.1082)--(1.7125,1.7709)--(2.8750,2.5809)--(5.2000,3.1455);
\fill[ink] (0.7825,0.5191) circle(1.65pt);
\fill[ink] (1.1313,1.1082) circle(1.65pt);
\fill[ink] (1.7125,1.7709) circle(1.65pt);
\fill[ink] (2.8750,2.5809) circle(1.65pt);
\fill[ink] (5.2000,3.1455) circle(1.65pt);
\draw[ink!65,thin] (5.24,3.1455)--(5.5200000000000005,3.6000);
\node[anchor=west,text=ink,font=\fontsize{7.5}{8.5}\selectfont] at (5.55,3.6000) {MT-AIME};
\node[anchor=west,font=\bfseries] at (7.6499999999999995,4.05) {(b) Multiple-choice scoring};
\draw[black!12] (7.6499999999999995,0.2)--(12.3,0.2);\node[anchor=east] at (7.529999999999999,0.2) {0};
\draw[black!12] (7.6499999999999995,1.01)--(12.3,1.01);\node[anchor=east] at (7.529999999999999,1.01) {2};
\draw[black!12] (7.6499999999999995,1.82)--(12.3,1.82);\node[anchor=east] at (7.529999999999999,1.82) {4};
\draw[black!12] (7.6499999999999995,2.6300000000000003)--(12.3,2.6300000000000003);\node[anchor=east] at (7.529999999999999,2.6300000000000003) {6};
\draw[black!12] (7.6499999999999995,3.4400000000000004)--(12.3,3.4400000000000004);\node[anchor=east] at (7.529999999999999,3.4400000000000004) {8};
\draw[black!50] (7.6499999999999995,.2)--(12.3,.2);
\node[anchor=north,font=\fontsize{7}{8}\selectfont] at (7.6499999999999995,.08) {0};
\node[anchor=north,font=\fontsize{7}{8}\selectfont] at (8.8125,.08) {500};
\node[anchor=north,font=\fontsize{7}{8}\selectfont] at (9.975,.08) {1000};
\node[anchor=north,font=\fontsize{7}{8}\selectfont] at (12.3,.08) {2000};
\node at (9.975,-.59) {Optimizer updates};
\node[rotate=90] at (7.09,1.85) {$\Delta$ accuracy (pp)};
\draw[forkblue,line width=.9pt] (7.6500,0.2000)--(7.8825,0.2000)--(8.2312,0.4045)--(8.8125,0.4045)--(9.9750,0.6091)--(12.3000,0.8136);
\fill[forkblue] (7.8825,0.2000) circle(1.65pt);
\fill[forkblue] (8.2312,0.4045) circle(1.65pt);
\fill[forkblue] (8.8125,0.4045) circle(1.65pt);
\fill[forkblue] (9.9750,0.6091) circle(1.65pt);
\fill[forkblue] (12.3000,0.8136) circle(1.65pt);
\draw[forkblue!65,thin] (12.34,0.8136)--(12.620000000000001,0.8136);
\node[anchor=west,text=forkblue,font=\fontsize{7.5}{8.5}\selectfont] at (12.65,0.8136) {GPQA-D};
\draw[forkteal,line width=.9pt] (7.6500,0.2000)--(7.8825,0.7063)--(8.2312,1.4150)--(8.8125,2.0731)--(9.9750,2.7313)--(12.3000,3.2375);
\fill[forkteal] (7.8825,0.7063) circle(1.65pt);
\fill[forkteal] (8.2312,1.4150) circle(1.65pt);
\fill[forkteal] (8.8125,2.0731) circle(1.65pt);
\fill[forkteal] (9.9750,2.7313) circle(1.65pt);
\fill[forkteal] (12.3000,3.2375) circle(1.65pt);
\draw[forkteal!65,thin] (12.34,3.2375)--(12.620000000000001,3.2375);
\node[anchor=west,text=forkteal,font=\fontsize{7.5}{8.5}\selectfont] at (12.65,3.2375) {MLogiQA};
\draw[forkorange,line width=.9pt] (7.6500,0.2000)--(7.8825,0.3805)--(8.2312,0.6212)--(8.8125,0.8620)--(9.9750,1.1629)--(12.3000,1.4036);
\fill[forkorange] (7.8825,0.3805) circle(1.65pt);
\fill[forkorange] (8.2312,0.6212) circle(1.65pt);
\fill[forkorange] (8.8125,0.8620) circle(1.65pt);
\fill[forkorange] (9.9750,1.1629) circle(1.65pt);
\fill[forkorange] (12.3000,1.4036) circle(1.65pt);
\draw[forkorange!65,thin] (12.34,1.4036)--(12.620000000000001,1.4036);
\node[anchor=west,text=forkorange,font=\fontsize{7.5}{8.5}\selectfont] at (12.65,1.4036) {CEval};
\draw[forkpurple,line width=.9pt,dashed] (7.6500,0.2000)--(7.8825,0.3356)--(8.2312,0.6874)--(8.8125,1.1691)--(9.9750,1.7430)--(12.3000,2.2276);
\fill[forkpurple] (7.8825,0.3356) circle(1.65pt);
\fill[forkpurple] (8.2312,0.6874) circle(1.65pt);
\fill[forkpurple] (8.8125,1.1691) circle(1.65pt);
\fill[forkpurple] (9.9750,1.7430) circle(1.65pt);
\fill[forkpurple] (12.3000,2.2276) circle(1.65pt);
\draw[forkpurple!65,thin] (12.34,2.2276)--(12.620000000000001,2.2276);
\node[anchor=west,text=forkpurple,font=\fontsize{7.5}{8.5}\selectfont] at (12.65,2.2276) {MMLU};
\end{tikzpicture}}
\caption{\textbf{The student continues to improve as training proceeds.}
Accuracy gains over the EDLM-4B base at $100$, $250$, $500$, $\num{1000}$,
and $\num{2000}$ updates, grouped by evaluation protocol on shared scales.
Every trajectory is nondecreasing; AIME25 and LMB-H plateau from $\num{1000}$
to $\num{2000}$ updates.}
\label{fig:training}
\end{figure}
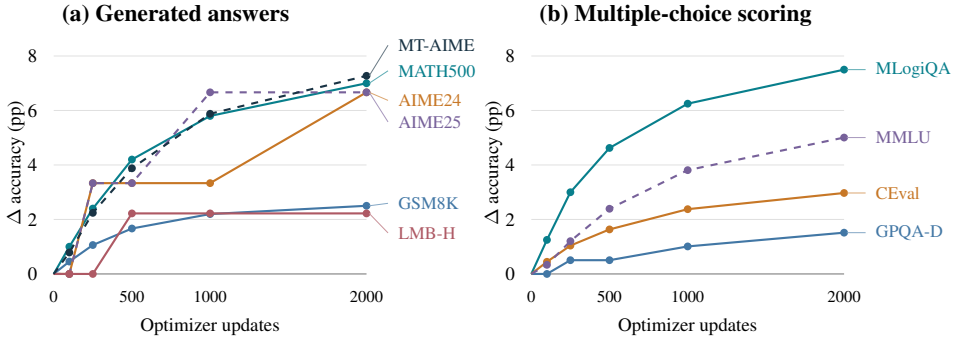

The gains develop throughout training (\autoref{fig:training}): every
benchmark is nondecreasing across the five evaluated checkpoints, by $500$
updates MATH500 reaches $76.80\%$ and MLogiQA $50.25\%$, and eight benchmarks
improve further between $\num{1000}$ and $\num{2000}$ updates while AIME25
and LMB-H remain level (counts in \autoref{tab:steps}).

\FloatBarrier
\subsection{Prefix alignment and entropy-first rollouts work together}
\label{sec:ablation}
The main results establish that the full method helps. We next test its
central design: should we change the order of student commitments, the
teacher--student information view, or both? We compare two rollout orders
(high or low entropy first) and two student views during OPD
(prefix-only or bidirectional). All four configurations generate training
responses with the native bidirectional student; a prefix-only loss changes
teaching, not the rollout.

\begin{figure}[!htb]
\centering
\resizebox{\linewidth}{!}{%
\begin{tikzpicture}[x=1cm,y=1cm,>=Latex,font=\fontsize{8}{9.3}\selectfont]
\path[use as bounding box] (0,-.5) rectangle (15.6,4.7);
\node[anchor=west,font=\bfseries] at (0,4.45) {(a) Mean gain over the EDLM-4B base (pp)};
\node[align=center,font=\fontsize{7.2}{8.3}\selectfont] at (3.35,3.37) {Low entropy first};
\node[align=center,font=\fontsize{7.2}{8.3}\selectfont] at (5.35,3.37) {High entropy first};
\node[anchor=east,align=right,font=\fontsize{7.2}{8.3}\selectfont] at (2.23,2.42) {Prefix-aligned\\view};
\fill[forkteal!29!white] (2.35,1.80) rectangle (4.35,3.05);
\draw[white,line width=1.2pt] (2.35,1.80) rectangle (4.35,3.05);
\node[text=ink,font=\bfseries\fontsize{11}{12}\selectfont] at (3.35,2.58) {+2.31};
\node[text=ink,font=\fontsize{6.6}{7.6}\selectfont] at (3.35,2.08) {9/10 up};
\fill[forkteal!62!white] (4.35,1.80) rectangle (6.35,3.05);
\draw[white,line width=1.2pt] (4.35,1.80) rectangle (6.35,3.05);
\node[text=white,font=\bfseries\fontsize{11}{12}\selectfont] at (5.35,2.58) {+4.93};
\node[text=white,font=\fontsize{6.6}{7.6}\selectfont] at (5.35,2.08) {10/10 up};
\node[anchor=east,align=right,font=\fontsize{7.2}{8.3}\selectfont] at (2.23,1.18) {Bidirectional\\view};
\fill[forkred!6!white] (2.35,0.55) rectangle (4.35,1.80);
\draw[white,line width=1.2pt] (2.35,0.55) rectangle (4.35,1.80);
\node[text=ink,font=\bfseries\fontsize{11}{12}\selectfont] at (3.35,1.33) {$-$0.50};
\node[text=ink,font=\fontsize{6.6}{7.6}\selectfont] at (3.35,0.83) {1/10 up};
\fill[forkteal!2!white] (4.35,0.55) rectangle (6.35,1.80);
\draw[white,line width=1.2pt] (4.35,0.55) rectangle (6.35,1.80);
\node[text=ink,font=\bfseries\fontsize{11}{12}\selectfont] at (5.35,1.33) {+0.14};
\node[text=ink,font=\fontsize{6.6}{7.6}\selectfont] at (5.35,0.83) {5/10 up};
\node[anchor=west,align=left,font=\fontsize{7.2}{8.3}\selectfont,text=ink!80] at (0,-.2) {View effect +3.80 pp \quad Order effect +1.63 pp \quad Interaction +1.99 pp};
\node[anchor=west,font=\bfseries] at (7.3,4.45) {(b) Each choice helps most with the other present};
\draw[ink!10] (8.25,0.55)--(12.75,0.55);
\node[anchor=east,font=\fontsize{6.6}{7.6}\selectfont,text=ink!70] at (8.18,0.55) {-4};
\draw[ink!10] (8.25,0.84)--(12.75,0.84);
\draw[ink!10] (8.25,1.13)--(12.75,1.13);
\node[anchor=east,font=\fontsize{6.6}{7.6}\selectfont,text=ink!70] at (8.18,1.13) {-2};
\draw[ink!10] (8.25,1.42)--(12.75,1.42);
\draw[ink!45] (8.25,1.71)--(12.75,1.71);
\node[anchor=east,font=\fontsize{6.6}{7.6}\selectfont,text=ink!70] at (8.18,1.71) {0};
\draw[ink!10] (8.25,2.00)--(12.75,2.00);
\draw[ink!10] (8.25,2.29)--(12.75,2.29);
\node[anchor=east,font=\fontsize{6.6}{7.6}\selectfont,text=ink!70] at (8.18,2.29) {2};
\draw[ink!10] (8.25,2.58)--(12.75,2.58);
\draw[ink!10] (8.25,2.87)--(12.75,2.87);
\node[anchor=east,font=\fontsize{6.6}{7.6}\selectfont,text=ink!70] at (8.18,2.87) {4};
\draw[ink!10] (8.25,3.16)--(12.75,3.16);
\draw[ink!10] (8.25,3.45)--(12.75,3.45);
\node[anchor=east,font=\fontsize{6.6}{7.6}\selectfont,text=ink!70] at (8.18,3.45) {6};
\draw[ink!10] (8.25,3.74)--(12.75,3.74);
\draw[ink!10] (8.25,4.03)--(12.75,4.03);
\node[anchor=east,font=\fontsize{6.6}{7.6}\selectfont,text=ink!70] at (8.18,4.03) {8};
\node[rotate=90,font=\fontsize{6.6}{7.6}\selectfont,text=ink!70] at (7.65,2.72) {Gain over base (pp)};
\node[anchor=north,font=\fontsize{7.2}{8.3}\selectfont] at (8.60,0.43) {Low entropy first};
\node[anchor=north,font=\fontsize{7.2}{8.3}\selectfont] at (12.40,0.43) {High entropy first};
\draw[forkteal!25,line width=.5pt] (8.60,2.02)--(12.40,2.44);
\draw[forkteal!25,line width=.5pt] (8.60,2.75)--(12.40,3.74);
\draw[forkteal!25,line width=.5pt] (8.60,2.68)--(12.40,3.64);
\draw[forkteal!25,line width=.5pt] (8.60,2.68)--(12.40,3.64);
\draw[forkteal!25,line width=.5pt] (8.60,1.71)--(12.40,2.35);
\draw[forkteal!25,line width=.5pt] (8.60,1.86)--(12.40,2.15);
\draw[forkteal!25,line width=.5pt] (8.60,2.76)--(12.40,3.88);
\draw[forkteal!25,line width=.5pt] (8.60,2.12)--(12.40,2.57);
\draw[forkteal!25,line width=.5pt] (8.60,2.37)--(12.40,3.16);
\draw[forkteal!25,line width=.5pt] (8.60,2.85)--(12.40,3.82);
\draw[forkpurple,line width=1.5pt] (8.60,1.57)--(12.40,1.75);
\fill[forkpurple] (8.52,1.49) rectangle (8.68,1.65);
\fill[forkpurple] (12.32,1.67) rectangle (12.48,1.83);
\node[anchor=west,text=forkpurple,font=\fontsize{7.2}{8.3}\selectfont] at (12.52,1.57) {+0.14};
\node[text=forkpurple,font=\bfseries\fontsize{7.2}{8.3}\selectfont,below] at (10.50,1.58) {Bidirectional view};
\draw[forkteal,line width=1.5pt] (8.60,2.38)--(12.40,3.14);
\fill[forkteal] (8.60,2.38) circle(2.4pt);
\fill[forkteal] (12.40,3.14) circle(2.4pt);
\node[anchor=west,text=forkteal,font=\fontsize{7.2}{8.3}\selectfont] at (12.52,3.32) {+4.93};
\node[text=forkteal,font=\bfseries\fontsize{7.2}{8.3}\selectfont,above] at (10.50,2.84) {Prefix-aligned view};
\node[anchor=west,font=\fontsize{6.6}{7.6}\selectfont,text=ink!70] at (7.65,-.2) {Bold: ten-benchmark mean. Faint: each benchmark under the prefix view.};
\end{tikzpicture}
}
\caption{\textbf{Entropy-first rollouts and prefix alignment reinforce each other.} (a) Mean gain over the EDLM-4B base for the four order--view configurations, with the number of benchmarks that improve. (b) The same means as an interaction plot; faint lines are the prefix-view benchmarks.}
\label{fig:ablation}
\end{figure}
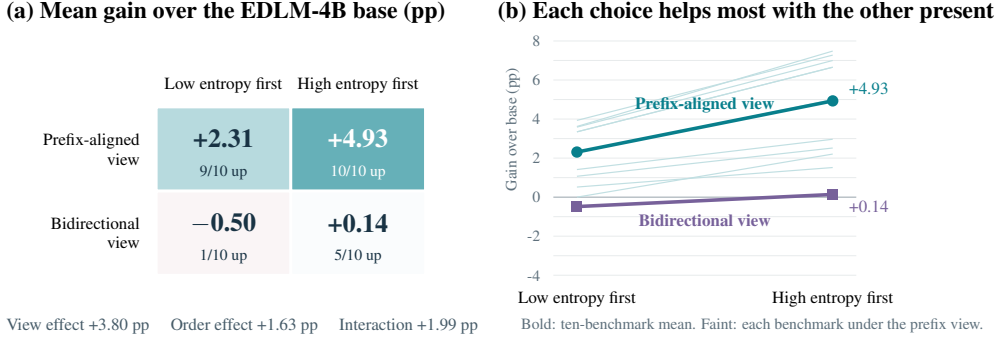

The proposed combination wins every benchmark (\autoref{fig:ablation}).
Under the prefix view, high-entropy-first training strictly improves all ten
tasks over low-entropy-first training; under high-entropy order,
prefix-aligned OPD strictly improves all ten tasks over bidirectional OPD
(per-benchmark counts in \autoref{tab:ablationcounts}).

The two changes reinforce each other. Averaged over the other factor, the
prefix view adds $3.80$ pp and entropy-first order $1.63$ pp, but the order
effect is $2.62$ pp under the prefix view and only $0.64$ pp under the
bidirectional view, a descriptive interaction of $1.99$ pp.

\FloatBarrier
\subsection{Teacher distributions add to SFT on the same questions}
\label{sec:subset}
\begin{figure}[!htb]
\centering
\resizebox{\linewidth}{!}{%
\begin{tikzpicture}[x=1cm,y=1cm,>=Latex,font=\fontsize{8}{9.3}\selectfont]
\path[use as bounding box] (0,-.6) rectangle (15.0,3.7);
\node[anchor=west,font=\bfseries] at (0.00,3.45) {(a) SFT$+$OPD minus SFT};
\fill[ink!8] (4.02,2.95) rectangle (4.04,3.15);
\node[anchor=east,font=\fontsize{7.2}{8.3}\selectfont] at (1.47,3.05) {MMLU};
\draw[forkteal,line width=1.6pt] (4.03,3.05)--(4.68,3.05);
\fill[forkteal] (4.68,3.05) circle(2.1pt);
\node[anchor=west,font=\fontsize{6.6}{7.6}\selectfont,text=forkteal] at (5.97,3.05) {+148 of 14\,042};
\fill[ink!8] (3.95,2.69) rectangle (4.11,2.89);
\node[anchor=east,font=\fontsize{7.2}{8.3}\selectfont] at (1.47,2.79) {MT-AIME};
\draw[forkteal,line width=1.6pt] (4.03,2.79)--(4.14,2.79);
\fill[forkteal] (4.14,2.79) circle(2.1pt);
\node[anchor=west,font=\fontsize{6.6}{7.6}\selectfont,text=forkteal] at (5.97,2.79) {+3 of 1\,650};
\fill[ink!8] (3.94,2.43) rectangle (4.12,2.63);
\node[anchor=east,font=\fontsize{7.2}{8.3}\selectfont] at (1.47,2.53) {CEval};
\draw[forkteal,line width=1.6pt] (4.03,2.53)--(4.21,2.53);
\fill[forkteal] (4.21,2.53) circle(2.1pt);
\node[anchor=west,font=\fontsize{6.6}{7.6}\selectfont,text=forkteal] at (5.97,2.53) {+4 of 1\,346};
\fill[ink!8] (3.94,2.17) rectangle (4.12,2.37);
\node[anchor=east,font=\fontsize{7.2}{8.3}\selectfont] at (1.47,2.27) {GSM8K};
\draw[forkteal,line width=1.6pt] (4.03,2.27)--(4.27,2.27);
\fill[forkteal] (4.27,2.27) circle(2.1pt);
\node[anchor=west,font=\fontsize{6.6}{7.6}\selectfont,text=forkteal] at (5.97,2.27) {+5 of 1\,319};
\fill[ink!8] (3.88,1.91) rectangle (4.19,2.11);
\node[anchor=east,font=\fontsize{7.2}{8.3}\selectfont] at (1.47,2.01) {MLogiQA};
\draw[forkteal,line width=1.6pt] (4.03,2.01)--(5.12,2.01);
\fill[forkteal] (5.12,2.01) circle(2.1pt);
\node[anchor=west,font=\fontsize{6.6}{7.6}\selectfont,text=forkteal] at (5.97,2.01) {+14 of 800};
\fill[ink!8] (3.78,1.65) rectangle (4.28,1.85);
\node[anchor=east,font=\fontsize{7.2}{8.3}\selectfont] at (1.47,1.75) {MATH500};
\draw[forkteal,line width=1.6pt] (4.03,1.75)--(4.65,1.75);
\fill[forkteal] (4.65,1.75) circle(2.1pt);
\node[anchor=west,font=\fontsize{6.6}{7.6}\selectfont,text=forkteal] at (5.97,1.75) {+5 of 500};
\fill[ink!8] (3.40,1.39) rectangle (4.66,1.59);
\node[anchor=east,font=\fontsize{7.2}{8.3}\selectfont] at (1.47,1.49) {GPQA-D};
\draw[forkred,line width=1.6pt] (4.03,1.49)--(3.09,1.49);
\fill[forkred] (3.09,1.49) circle(2.1pt);
\node[anchor=west,font=\fontsize{6.6}{7.6}\selectfont,text=forkred] at (5.97,1.49) {$-$3 of 198};
\fill[ink!8] (1.55,1.13) rectangle (5.89,1.33);
\node[anchor=east,font=\fontsize{7.2}{8.3}\selectfont] at (1.47,1.23) {LMB-H};
\draw[ink!45,line width=1.6pt] (4.03,1.23)--(2.65,1.23);
\fill[ink!45] (2.65,1.23) circle(2.1pt);
\node[anchor=west,font=\fontsize{6.6}{7.6}\selectfont,text=ink!60] at (5.97,1.23) {$-$1 of 45};
\fill[ink!8] (1.55,0.87) rectangle (5.89,1.07);
\node[anchor=east,font=\fontsize{7.2}{8.3}\selectfont] at (1.47,0.97) {AIME24};
\draw[ink!45,line width=1.6pt] (4.03,0.97)--(1.96,0.97);
\fill[ink!45] (1.96,0.97) circle(2.1pt);
\node[anchor=west,font=\fontsize{6.6}{7.6}\selectfont,text=ink!60] at (5.97,0.97) {$-$1 of 30};
\fill[ink!8] (1.55,0.61) rectangle (5.89,0.81);
\node[anchor=east,font=\fontsize{7.2}{8.3}\selectfont] at (1.47,0.71) {AIME25};
\draw[ink!45,line width=1.6pt] (4.03,0.71)--(1.96,0.71);
\fill[ink!45] (1.96,0.71) circle(2.1pt);
\node[anchor=west,font=\fontsize{6.6}{7.6}\selectfont,text=ink!60] at (5.97,0.71) {$-$1 of 30};
\draw[ink!45] (4.03,0.56)--(4.03,3.18);
\draw[ink!45] (1.55,0.56)--(1.55,0.48);
\node[anchor=north,font=\fontsize{6.6}{7.6}\selectfont,text=ink!70] at (1.55,0.46) {$-$4};
\draw[ink!45] (2.17,0.56)--(2.17,0.48);
\node[anchor=north,font=\fontsize{6.6}{7.6}\selectfont,text=ink!70] at (2.17,0.46) {$-$3};
\draw[ink!45] (2.79,0.56)--(2.79,0.48);
\node[anchor=north,font=\fontsize{6.6}{7.6}\selectfont,text=ink!70] at (2.79,0.46) {$-$2};
\draw[ink!45] (3.41,0.56)--(3.41,0.48);
\node[anchor=north,font=\fontsize{6.6}{7.6}\selectfont,text=ink!70] at (3.41,0.46) {$-$1};
\draw[ink!45] (4.03,0.56)--(4.03,0.48);
\node[anchor=north,font=\fontsize{6.6}{7.6}\selectfont,text=ink!70] at (4.03,0.46) {0};
\draw[ink!45] (4.65,0.56)--(4.65,0.48);
\node[anchor=north,font=\fontsize{6.6}{7.6}\selectfont,text=ink!70] at (4.65,0.46) {+1};
\draw[ink!45] (5.27,0.56)--(5.27,0.48);
\node[anchor=north,font=\fontsize{6.6}{7.6}\selectfont,text=ink!70] at (5.27,0.46) {+2};
\draw[ink!45] (5.89,0.56)--(5.89,0.48);
\node[anchor=north,font=\fontsize{6.6}{7.6}\selectfont,text=ink!70] at (5.89,0.46) {+3};
\node[anchor=north,font=\fontsize{7.2}{8.3}\selectfont,text=ink!75] at (3.72,0.22) {Accuracy change (pp); grey: within two items};
\node[anchor=west,font=\bfseries] at (7.20,3.45) {(b) SFT$+$OPD minus OPD};
\fill[ink!8] (11.22,2.95) rectangle (11.24,3.15);
\node[anchor=east,font=\fontsize{7.2}{8.3}\selectfont] at (8.67,3.05) {MMLU};
\draw[forkteal,line width=1.6pt] (11.23,3.05)--(12.21,3.05);
\fill[forkteal] (12.21,3.05) circle(2.1pt);
\node[anchor=west,font=\fontsize{6.6}{7.6}\selectfont,text=forkteal] at (13.17,3.05) {+222 of 14\,042};
\fill[ink!8] (11.15,2.69) rectangle (11.31,2.89);
\node[anchor=east,font=\fontsize{7.2}{8.3}\selectfont] at (8.67,2.79) {MT-AIME};
\draw[forkteal,line width=1.6pt] (11.23,2.79)--(11.53,2.79);
\fill[forkteal] (11.53,2.79) circle(2.1pt);
\node[anchor=west,font=\fontsize{6.6}{7.6}\selectfont,text=forkteal] at (13.17,2.79) {+8 of 1\,650};
\fill[ink!8] (11.14,2.43) rectangle (11.32,2.63);
\node[anchor=east,font=\fontsize{7.2}{8.3}\selectfont] at (8.67,2.53) {CEval};
\draw[forkteal,line width=1.6pt] (11.23,2.53)--(11.64,2.53);
\fill[forkteal] (11.64,2.53) circle(2.1pt);
\node[anchor=west,font=\fontsize{6.6}{7.6}\selectfont,text=forkteal] at (13.17,2.53) {+9 of 1\,346};
\fill[ink!8] (11.14,2.17) rectangle (11.32,2.37);
\node[anchor=east,font=\fontsize{7.2}{8.3}\selectfont] at (8.67,2.27) {GSM8K};
\draw[forkteal,line width=1.6pt] (11.23,2.27)--(11.75,2.27);
\fill[forkteal] (11.75,2.27) circle(2.1pt);
\node[anchor=west,font=\fontsize{6.6}{7.6}\selectfont,text=forkteal] at (13.17,2.27) {+11 of 1\,319};
\fill[ink!8] (11.08,1.91) rectangle (11.38,2.11);
\node[anchor=east,font=\fontsize{7.2}{8.3}\selectfont] at (8.67,2.01) {MLogiQA};
\draw[forkteal,line width=1.6pt] (11.23,2.01)--(12.62,2.01);
\fill[forkteal] (12.62,2.01) circle(2.1pt);
\node[anchor=west,font=\fontsize{6.6}{7.6}\selectfont,text=forkteal] at (13.17,2.01) {+18 of 800};
\fill[ink!8] (10.98,1.65) rectangle (11.48,1.85);
\node[anchor=east,font=\fontsize{7.2}{8.3}\selectfont] at (8.67,1.75) {MATH500};
\draw[forkteal,line width=1.6pt] (11.23,1.75)--(12.22,1.75);
\fill[forkteal] (12.22,1.75) circle(2.1pt);
\node[anchor=west,font=\fontsize{6.6}{7.6}\selectfont,text=forkteal] at (13.17,1.75) {+8 of 500};
\fill[ink!8] (10.60,1.39) rectangle (11.86,1.59);
\node[anchor=east,font=\fontsize{7.2}{8.3}\selectfont] at (8.67,1.49) {GPQA-D};
\fill[ink!45] (11.23,1.49) circle(2.1pt);
\node[anchor=west,font=\fontsize{6.6}{7.6}\selectfont,text=ink!60] at (13.17,1.49) {0 of 198};
\fill[ink!8] (8.75,1.13) rectangle (13.09,1.33);
\node[anchor=east,font=\fontsize{7.2}{8.3}\selectfont] at (8.67,1.23) {LMB-H};
\fill[ink!45] (11.23,1.23) circle(2.1pt);
\node[anchor=west,font=\fontsize{6.6}{7.6}\selectfont,text=ink!60] at (13.17,1.23) {0 of 45};
\fill[ink!8] (8.75,0.87) rectangle (13.09,1.07);
\node[anchor=east,font=\fontsize{7.2}{8.3}\selectfont] at (8.67,0.97) {AIME24};
\fill[ink!45] (11.23,0.97) circle(2.1pt);
\node[anchor=west,font=\fontsize{6.6}{7.6}\selectfont,text=ink!60] at (13.17,0.97) {0 of 30};
\fill[ink!8] (8.75,0.61) rectangle (13.09,0.81);
\node[anchor=east,font=\fontsize{7.2}{8.3}\selectfont] at (8.67,0.71) {AIME25};
\fill[ink!45] (11.23,0.71) circle(2.1pt);
\node[anchor=west,font=\fontsize{6.6}{7.6}\selectfont,text=ink!60] at (13.17,0.71) {0 of 30};
\draw[ink!45] (11.23,0.56)--(11.23,3.18);
\draw[ink!45] (8.75,0.56)--(8.75,0.48);
\node[anchor=north,font=\fontsize{6.6}{7.6}\selectfont,text=ink!70] at (8.75,0.46) {$-$4};
\draw[ink!45] (9.37,0.56)--(9.37,0.48);
\node[anchor=north,font=\fontsize{6.6}{7.6}\selectfont,text=ink!70] at (9.37,0.46) {$-$3};
\draw[ink!45] (9.99,0.56)--(9.99,0.48);
\node[anchor=north,font=\fontsize{6.6}{7.6}\selectfont,text=ink!70] at (9.99,0.46) {$-$2};
\draw[ink!45] (10.61,0.56)--(10.61,0.48);
\node[anchor=north,font=\fontsize{6.6}{7.6}\selectfont,text=ink!70] at (10.61,0.46) {$-$1};
\draw[ink!45] (11.23,0.56)--(11.23,0.48);
\node[anchor=north,font=\fontsize{6.6}{7.6}\selectfont,text=ink!70] at (11.23,0.46) {0};
\draw[ink!45] (11.85,0.56)--(11.85,0.48);
\node[anchor=north,font=\fontsize{6.6}{7.6}\selectfont,text=ink!70] at (11.85,0.46) {+1};
\draw[ink!45] (12.47,0.56)--(12.47,0.48);
\node[anchor=north,font=\fontsize{6.6}{7.6}\selectfont,text=ink!70] at (12.47,0.46) {+2};
\draw[ink!45] (13.09,0.56)--(13.09,0.48);
\node[anchor=north,font=\fontsize{6.6}{7.6}\selectfont,text=ink!70] at (13.09,0.46) {+3};
\node[anchor=north,font=\fontsize{7.2}{8.3}\selectfont,text=ink!75] at (10.92,0.22) {Accuracy change (pp); grey: within two items};
\end{tikzpicture}
}
\caption{\textbf{Teacher distributions add to hard targets on the same questions.} All three objectives train on the teacher-correct, student-wrong questions of the main run; rows are ordered by benchmark size, right labels give the net change in correct answers, and grey marks changes of at most two items. (a) Adding OPD to SFT raises six benchmarks; GPQA-D is the only larger decrease. (b) Adding SFT to OPD lowers none.}
\label{fig:subset}
\end{figure}
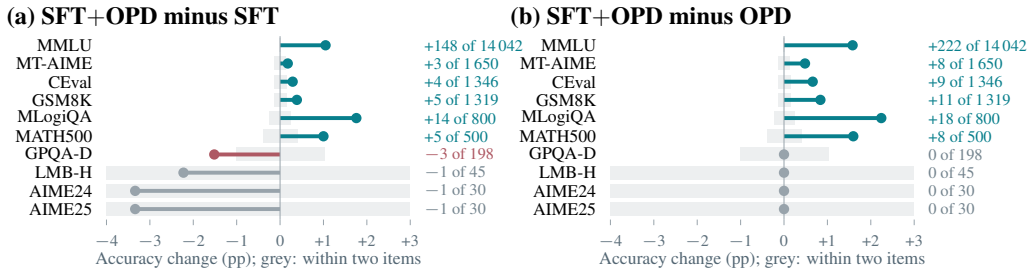

Part of a distillation gain could come from practicing more correct answers
rather than from the teacher's distribution. The teacher can add information
only where it knows something the student does not, namely on questions it
answers correctly and the student does not; in the main run the routing
places about $30\%$ of the training questions on this route. We train SFT,
OPD, and SFT+OPD on exactly these questions, so the three objectives differ
only in the supervision they apply. All three improve the base on nine or ten
benchmarks, and SFT and OPD are close on most of them.

We draw two points from \autoref{fig:subset}. First, the objectives are
complementary: adding OPD to SFT improves six benchmarks, by $148$ net
correct answers on MMLU, $14$ on MLogiQA, $5$ each on GSM8K and MATH500, $4$
on CEval, and $3$ on MT-AIME, and adding SFT to OPD lowers none. AIME24,
AIME25, and LMB-H move by a single item each, and GPQA-D, three items below
SFT, is the only larger decrease. Second, the two
signals differ in kind: SFT places its target on one token, whereas OPD
supplies the teacher's distribution over alternatives after the student's own
prefix. For a shared logit vector the gradients are $p-e_y$ and $p-t$, and any
mixture of the two objectives targets the interpolated distribution
(\autoref{app:objectives}), so the combination adds the teacher's ranking of
alternatives to the hard target rather than repeating it.

\FloatBarrier
\subsection{Gains rise with teacher strength for both student sizes}
\label{sec:scale}
\begin{figure}[!htb]
\centering
\resizebox{\linewidth}{!}{%
\begin{tikzpicture}[x=1cm,y=1cm,>=Latex,font=\fontsize{8}{9.3}\selectfont]
\path[use as bounding box] (0,-.6) rectangle (14.5,3.95);
\node[anchor=west,font=\bfseries] at (0,3.72) {(a) Mean gain over each student's own base (pp)};
\node[align=center,font=\fontsize{7.2}{8.3}\selectfont] at (2.50,3.33) {Q3-4B};
\node[align=center,font=\fontsize{6.6}{7.6}\selectfont,text=ink!65] at (2.50,3.05) {47.0};
\node[align=center,font=\fontsize{7.2}{8.3}\selectfont] at (3.80,3.33) {Q2.5-7B};
\node[align=center,font=\fontsize{6.6}{7.6}\selectfont,text=ink!65] at (3.80,3.05) {47.2};
\node[align=center,font=\fontsize{7.2}{8.3}\selectfont] at (5.10,3.33) {Q3-8B};
\node[align=center,font=\fontsize{6.6}{7.6}\selectfont,text=ink!65] at (5.10,3.05) {49.9};
\node[align=center,font=\fontsize{7.2}{8.3}\selectfont] at (6.40,3.33) {Q3-30B};
\node[align=center,font=\fontsize{6.6}{7.6}\selectfont,text=ink!65] at (6.40,3.05) {53.5};
\node[anchor=east,font=\fontsize{7.2}{8.3}\selectfont] at (1.73,2.39) {EDLM-4B};
\fill[forkteal!36!white] (1.85,1.91) rectangle (3.15,2.87);
\draw[white,line width=1.2pt] (1.85,1.91) rectangle (3.15,2.87);
\node[text=ink,font=\bfseries\fontsize{10.5}{11.5}\selectfont] at (2.50,2.55) {+2.91};
\node[text=ink,font=\fontsize{6.6}{7.6}\selectfont] at (2.50,2.07) {10/10 up};
\fill[forkteal!40!white] (3.15,1.91) rectangle (4.45,2.87);
\draw[white,line width=1.2pt] (3.15,1.91) rectangle (4.45,2.87);
\node[text=ink,font=\bfseries\fontsize{10.5}{11.5}\selectfont] at (3.80,2.55) {+3.17};
\node[text=ink,font=\fontsize{6.6}{7.6}\selectfont] at (3.80,2.07) {10/10 up};
\fill[forkteal!44!white] (4.45,1.91) rectangle (5.75,2.87);
\draw[white,line width=1.2pt] (4.45,1.91) rectangle (5.75,2.87);
\node[text=ink,font=\bfseries\fontsize{10.5}{11.5}\selectfont] at (5.10,2.55) {+3.51};
\node[text=ink,font=\fontsize{6.6}{7.6}\selectfont] at (5.10,2.07) {10/10 up};
\fill[forkteal!62!white] (5.75,1.91) rectangle (7.05,2.87);
\draw[white,line width=1.2pt] (5.75,1.91) rectangle (7.05,2.87);
\node[text=white,font=\bfseries\fontsize{10.5}{11.5}\selectfont] at (6.40,2.55) {+4.93};
\node[text=white,font=\fontsize{6.6}{7.6}\selectfont] at (6.40,2.07) {10/10 up};
\node[anchor=east,font=\fontsize{7.2}{8.3}\selectfont] at (1.73,1.43) {EDLM-8B};
\fill[pattern=north east lines,pattern color=ink!25] (1.85,0.95) rectangle (3.15,1.91);
\draw[white,line width=1.2pt] (1.85,0.95) rectangle (3.15,1.91);
\fill[pattern=north east lines,pattern color=ink!25] (3.15,0.95) rectangle (4.45,1.91);
\draw[white,line width=1.2pt] (3.15,0.95) rectangle (4.45,1.91);
\fill[forkteal!19!white] (4.45,0.95) rectangle (5.75,1.91);
\draw[white,line width=1.2pt] (4.45,0.95) rectangle (5.75,1.91);
\node[text=ink,font=\bfseries\fontsize{10.5}{11.5}\selectfont] at (5.10,1.59) {+1.50};
\node[text=ink,font=\fontsize{6.6}{7.6}\selectfont] at (5.10,1.11) {9/10 up};
\fill[forkteal!40!white] (5.75,0.95) rectangle (7.05,1.91);
\draw[white,line width=1.2pt] (5.75,0.95) rectangle (7.05,1.91);
\node[text=ink,font=\bfseries\fontsize{10.5}{11.5}\selectfont] at (6.40,1.59) {+3.22};
\node[text=ink,font=\fontsize{6.6}{7.6}\selectfont] at (6.40,1.11) {9/10 up};
\node[anchor=west,align=left,font=\fontsize{6.6}{7.6}\selectfont,text=ink!70,text width=6.9cm] at (0,0.45) {Q3 = Qwen3, Q2.5 = Qwen2.5; the 30B teacher is Qwen3-30B-A3B. The number under each teacher is its strength: mean accuracy on the nine benchmarks all four report. Hatched: not run.};
\node[anchor=west,font=\bfseries] at (7.3,3.72) {(b) Gain against teacher strength};
\draw[ink!10] (8.35,0.60)--(13.20,0.60);
\draw[ink!10] (8.35,0.80)--(13.20,0.80);
\node[anchor=east,font=\fontsize{6.6}{7.6}\selectfont,text=ink!70] at (8.28,0.80) {-4};
\draw[ink!10] (8.35,1.01)--(13.20,1.01);
\draw[ink!10] (8.35,1.21)--(13.20,1.21);
\node[anchor=east,font=\fontsize{6.6}{7.6}\selectfont,text=ink!70] at (8.28,1.21) {-2};
\draw[ink!10] (8.35,1.42)--(13.20,1.42);
\draw[ink!45] (8.35,1.62)--(13.20,1.62);
\node[anchor=east,font=\fontsize{6.6}{7.6}\selectfont,text=ink!70] at (8.28,1.62) {0};
\draw[ink!10] (8.35,1.82)--(13.20,1.82);
\draw[ink!10] (8.35,2.03)--(13.20,2.03);
\node[anchor=east,font=\fontsize{6.6}{7.6}\selectfont,text=ink!70] at (8.28,2.03) {2};
\draw[ink!10] (8.35,2.23)--(13.20,2.23);
\draw[ink!10] (8.35,2.43)--(13.20,2.43);
\node[anchor=east,font=\fontsize{6.6}{7.6}\selectfont,text=ink!70] at (8.28,2.43) {4};
\draw[ink!10] (8.35,2.64)--(13.20,2.64);
\draw[ink!10] (8.35,2.84)--(13.20,2.84);
\node[anchor=east,font=\fontsize{6.6}{7.6}\selectfont,text=ink!70] at (8.28,2.84) {6};
\draw[ink!10] (8.35,3.05)--(13.20,3.05);
\draw[ink!10] (8.35,3.25)--(13.20,3.25);
\node[anchor=east,font=\fontsize{6.6}{7.6}\selectfont,text=ink!70] at (8.28,3.25) {8};
\node[rotate=90,font=\fontsize{6.6}{7.6}\selectfont,text=ink!70] at (7.75,2.13) {Gain over base (pp)};
\draw[ink!35] (8.55,0.58)--(8.55,0.50);
\node[anchor=north,font=\fontsize{6.6}{7.6}\selectfont,text=ink!70] at (8.55,0.48) {46};
\draw[ink!35] (9.60,0.58)--(9.60,0.50);
\node[anchor=north,font=\fontsize{6.6}{7.6}\selectfont,text=ink!70] at (9.60,0.48) {48};
\draw[ink!35] (10.64,0.58)--(10.64,0.50);
\node[anchor=north,font=\fontsize{6.6}{7.6}\selectfont,text=ink!70] at (10.64,0.48) {50};
\draw[ink!35] (11.69,0.58)--(11.69,0.50);
\node[anchor=north,font=\fontsize{6.6}{7.6}\selectfont,text=ink!70] at (11.69,0.48) {52};
\draw[ink!35] (12.74,0.58)--(12.74,0.50);
\node[anchor=north,font=\fontsize{6.6}{7.6}\selectfont,text=ink!70] at (12.74,0.48) {54};
\node[anchor=north,font=\fontsize{6.6}{7.6}\selectfont,text=ink!70] at (10.78,0.22) {Teacher strength (mean accuracy over the nine shared benchmarks, \%)};
\draw[forkteal!22,line width=.45pt] (9.09,1.90) -- (10.60,2.04) -- (12.50,2.13);
\draw[forkteal!22,line width=.45pt] (9.09,2.39) -- (10.60,2.68) -- (12.50,3.05);
\draw[forkteal!22,line width=.45pt] (9.09,2.30) -- (10.60,2.30) -- (12.50,2.98);
\draw[forkteal!22,line width=.45pt] (9.09,2.30) -- (10.60,2.30) -- (12.50,2.98);
\draw[forkteal!22,line width=.45pt] (9.09,2.07) -- (10.60,2.07) -- (12.50,2.07);
\draw[forkteal!22,line width=.45pt] (9.09,1.72) -- (10.60,1.83) -- (12.50,1.93);
\draw[forkteal!22,line width=.45pt] (9.09,2.54) -- (10.60,2.82) -- (12.50,3.15);
\draw[forkteal!22,line width=.45pt] (9.09,2.00) -- (10.60,2.12) -- (12.50,2.23);
\draw[forkteal!22,line width=.45pt] (9.09,2.23) -- (10.60,2.40) -- (12.50,2.64);
\draw[forkteal!22,line width=.45pt] (9.09,2.68) -- (10.60,2.79) -- (12.50,3.10);
\draw[forkteal,line width=1.5pt] (9.09,2.21) -- (10.60,2.33) -- (12.50,2.62);
\fill[forkteal] (9.09,2.21) circle(2.3pt);
\node[anchor=south,text=forkteal,font=\fontsize{7.2}{8.3}\selectfont] at (9.09,2.30) {+2.91};
\fill[forkteal] (10.60,2.33) circle(2.3pt);
\node[anchor=south,text=forkteal,font=\fontsize{7.2}{8.3}\selectfont] at (10.60,2.42) {+3.51};
\fill[forkteal] (12.50,2.62) circle(2.3pt);
\node[anchor=south,text=forkteal,font=\fontsize{7.2}{8.3}\selectfont] at (12.50,2.71) {+4.93};
\node[anchor=west,text=forkteal,font=\bfseries\fontsize{7.2}{8.3}\selectfont] at (12.72,2.62) {EDLM-4B};
\draw[forkblue,line width=1.5pt] (10.60,1.92) -- (12.50,2.28);
\fill[forkblue] (10.52,1.84) rectangle (10.68,2.00);
\node[anchor=north,text=forkblue,font=\fontsize{7.2}{8.3}\selectfont] at (10.60,1.83) {+1.50};
\fill[forkblue] (12.42,2.20) rectangle (12.58,2.36);
\node[anchor=north,text=forkblue,font=\fontsize{7.2}{8.3}\selectfont] at (12.50,2.19) {+3.22};
\node[anchor=west,text=forkblue,font=\bfseries\fontsize{7.2}{8.3}\selectfont] at (12.72,2.28) {EDLM-8B};
\draw[forkteal,line width=1pt,fill=white] (9.17,2.27) circle(2.3pt);
\node[anchor=north,text=forkteal,font=\fontsize{6.6}{7.6}\selectfont,inner sep=1pt,fill=white] at (9.37,2.07) {Q2.5-7B};
\end{tikzpicture}
}
\caption{\textbf{Gains rise with teacher strength for both student sizes.} (a) Mean gain over each student's own base for every trained pair, with the number of benchmarks that improve. (b) The same means against teacher strength; faint lines are the EDLM-4B benchmarks, the hollow marker is the Qwen2.5-7B teacher, and no curve is fitted.}
\label{fig:scale}
\end{figure}
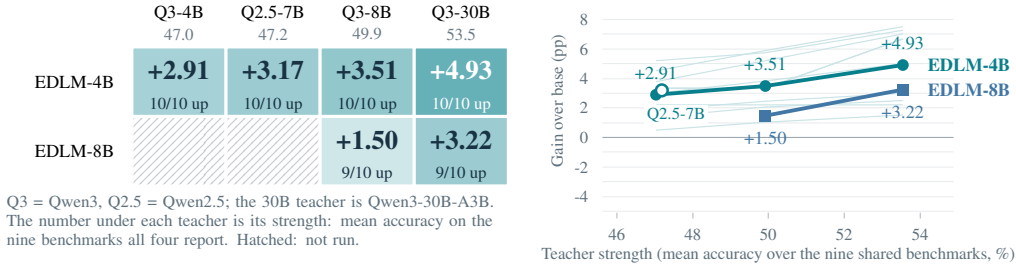

If ForkLeft transfers teacher information, a stronger teacher should produce
a larger gain. We test Qwen3 teachers at 4B, 8B, and 30B-A3B and
Qwen2.5-7B-Base for EDLM-4B, and Qwen3 at 8B and 30B-A3B for EDLM-8B, with
gains relative to each student's own base; because the 30B teacher activates
only about 3B parameters per token, we order teachers by strength, their mean
accuracy on the nine benchmarks all four report.

The trend holds for both students (\autoref{fig:scale}). For EDLM-4B the
ten-task mean gain rises from $2.91$ pp with Qwen3-4B through $3.17$ with
Qwen2.5-7B and $3.51$ with Qwen3-8B to $4.93$ pp with Qwen3-30B, and the
30B teacher beats the 8B teacher on nine benchmarks and ties LMB-H. The 8B
student, despite its stronger base, follows the same trend: its mean gain
rises from $1.50$ to $3.22$ pp, and the larger teacher wins all ten benchmark
comparisons against the 8B teacher; relative to the base, nine benchmarks
improve and LMB-H changes from $8/45$ to $7/45$.

The rise is an empirical scaling trend: within the tested range, every
increase in teacher strength increased the distilled gain for both student
sizes, and the trend had not flattened at the strongest teacher. The points
are actual checkpoints, including a mixture-of-experts teacher, so we fit no
law (counts in \autoref{tab:scalecounts}).

\FloatBarrier
\subsection{The recipe generalizes to a second student family}
\label{sec:generalize}
\begin{figure}[!htb]
\centering
\resizebox{\linewidth}{!}{%
\begin{tikzpicture}[x=1cm,y=1cm,>=Latex,font=\fontsize{8}{9.3}\selectfont]
\path[use as bounding box] (0,0) rectangle (14,1.12);
\node[anchor=south,font=\fontsize{7.2}{8.3}\selectfont] at (2.30,0.74) {GSM8K};
\node[anchor=south,font=\fontsize{7.2}{8.3}\selectfont] at (3.40,0.74) {MATH500};
\node[anchor=south,font=\fontsize{7.2}{8.3}\selectfont] at (4.50,0.74) {AIME24};
\node[anchor=south,font=\fontsize{7.2}{8.3}\selectfont] at (5.60,0.74) {AIME25};
\node[anchor=south,font=\fontsize{7.2}{8.3}\selectfont] at (6.70,0.74) {LMB-H};
\node[anchor=south,font=\fontsize{7.2}{8.3}\selectfont] at (7.80,0.74) {GPQA-D};
\node[anchor=south,font=\fontsize{7.2}{8.3}\selectfont] at (8.90,0.74) {MLogiQA};
\node[anchor=south,font=\fontsize{7.2}{8.3}\selectfont] at (10.00,0.74) {CEval};
\node[anchor=south,font=\fontsize{7.2}{8.3}\selectfont] at (11.10,0.74) {MMLU};
\node[anchor=south,font=\fontsize{7.2}{8.3}\selectfont] at (12.20,0.74) {MT-AIME};
\node[anchor=east,font=\bfseries\fontsize{7.2}{8.3}\selectfont] at (1.63,0.37) {SDAR-4B};
\fill[forkteal!24!white] (1.75,0.04) rectangle (2.85,0.70);
\draw[white,line width=1.2pt] (1.75,0.04) rectangle (2.85,0.70);
\node[text=ink,font=\bfseries\fontsize{7.2}{8.3}\selectfont] at (2.30,0.49) {+1.90};
\node[text=ink,font=\fontsize{6.6}{7.6}\selectfont] at (2.30,0.18) {+25 items};
\fill[forkteal!12!white] (2.85,0.04) rectangle (3.95,0.70);
\draw[white,line width=1.2pt] (2.85,0.04) rectangle (3.95,0.70);
\node[text=ink,font=\bfseries\fontsize{7.2}{8.3}\selectfont] at (3.40,0.49) {+1.00};
\node[text=ink,font=\fontsize{6.6}{7.6}\selectfont] at (3.40,0.18) {+5 items};
\fill[black!7] (3.95,0.04) rectangle (5.05,0.70);
\draw[white,line width=1.2pt] (3.95,0.04) rectangle (5.05,0.70);
\node[text=ink!75,font=\fontsize{7.2}{8.3}\selectfont] at (4.50,0.37) {$-$2 items};
\fill[black!7] (5.05,0.04) rectangle (6.15,0.70);
\draw[white,line width=1.2pt] (5.05,0.04) rectangle (6.15,0.70);
\node[text=ink!75,font=\fontsize{7.2}{8.3}\selectfont] at (5.60,0.37) {0 items};
\fill[black!7] (6.15,0.04) rectangle (7.25,0.70);
\draw[white,line width=1.2pt] (6.15,0.04) rectangle (7.25,0.70);
\node[text=ink!75,font=\fontsize{7.2}{8.3}\selectfont] at (6.70,0.37) {+1 item};
\fill[forkteal!57!white] (7.25,0.04) rectangle (8.35,0.70);
\draw[white,line width=1.2pt] (7.25,0.04) rectangle (8.35,0.70);
\node[text=ink,font=\bfseries\fontsize{7.2}{8.3}\selectfont] at (7.80,0.49) {+4.55};
\node[text=ink,font=\fontsize{6.6}{7.6}\selectfont] at (7.80,0.18) {+9 items};
\fill[black!7] (8.35,0.04) rectangle (9.45,0.70);
\draw[white,line width=1.2pt] (8.35,0.04) rectangle (9.45,0.70);
\node[text=ink!75,font=\fontsize{7.2}{8.3}\selectfont] at (8.90,0.37) {$-$1 item};
\fill[forkteal!35!white] (9.45,0.04) rectangle (10.55,0.70);
\draw[white,line width=1.2pt] (9.45,0.04) rectangle (10.55,0.70);
\node[text=ink,font=\bfseries\fontsize{7.2}{8.3}\selectfont] at (10.00,0.49) {+2.82};
\node[text=ink,font=\fontsize{6.6}{7.6}\selectfont] at (10.00,0.18) {+38 items};
\fill[forkteal!88!white] (10.55,0.04) rectangle (11.65,0.70);
\draw[white,line width=1.2pt] (10.55,0.04) rectangle (11.65,0.70);
\node[text=white,font=\bfseries\fontsize{7.2}{8.3}\selectfont] at (11.10,0.49) {+7.06};
\node[text=white,font=\fontsize{6.6}{7.6}\selectfont] at (11.10,0.18) {+991 items};
\fill[forkteal!5!white] (11.65,0.04) rectangle (12.75,0.70);
\draw[white,line width=1.2pt] (11.65,0.04) rectangle (12.75,0.70);
\node[text=ink,font=\bfseries\fontsize{7.2}{8.3}\selectfont] at (12.20,0.49) {+0.42};
\node[text=ink,font=\fontsize{6.6}{7.6}\selectfont] at (12.20,0.18) {+7 items};
\node[anchor=west,font=\bfseries\fontsize{7.2}{8.3}\selectfont] at (12.87,0.37) {7/10 up};
\end{tikzpicture}
}
\caption{\textbf{ForkLeft generalizes to SDAR-4B.} Accuracy change per benchmark after $500$ updates with Qwen3-30B-A3B-Base (pp and net correct answers); grey: within two items.}
\label{fig:transfer}
\end{figure}
We ask next whether the recipe helps a diffusion family it was not developed
on, and distill the same Qwen3-30B teacher into SDAR-4B \citep{sdar2026}, a
block-diffusion model converted from Qwen3-4B, for $500$ updates. Seven
benchmarks improve, including MMLU by $7.06$ pp and GPQA-D by $4.55$ pp, and
the other three move by at most two items (\autoref{fig:transfer},
\autoref{tab:sdarcounts}).

\FloatBarrier
\subsection{Comparison with published diffusion models at matched scale}
\label{sec:external}

\begin{table}[!htb]
\centering
\caption{\textbf{Distilled students exceed published diffusion models of the same size on seven benchmarks.} Accuracy (\%); SDAR-Chat and OPDLM scores from \citet{opdlm2026} under our prompt template and scorer.}
\label{tab:external}
{\footnotesize\setlength{\tabcolsep}{4.5pt}\renewcommand{\arraystretch}{1.0}
\begin{tabular}{lrrrrrrr}
\toprule
Model & MATH500 & AIME24 & AIME25 & MLogiQA & CEval & MMLU & MT-AIME\\
\midrule
SDAR-Chat-4B & 72.80 & 10.00 & 7.50 & 46.50 & 62.90 & 74.90 & 3.00\\
OPDLM-4B & 72.80 & 14.40 & 12.60 & 46.50 & 66.90 & 65.50 & 5.30\\
EDLM-4B $+$ ForkLeft & \textbf{79.60} & \textbf{16.67} & \textbf{13.33} & \textbf{53.13} & \textbf{73.70} & \textbf{76.05} & \textbf{15.82}\\
\midrule
SDAR-Chat-8B & 78.60 & 10.00 & 10.00 & 46.30 & 70.20 & 78.60 & 4.00\\
OPDLM-8B & 71.20 & 14.70 & 12.40 & 42.00 & 73.30 & 70.90 & 7.90\\
EDLM-8B $+$ ForkLeft & \textbf{81.20} & \textbf{20.00} & \textbf{16.67} & \textbf{55.38} & \textbf{78.08} & \textbf{79.11} & \textbf{18.00}\\
\bottomrule
\end{tabular}}
\end{table}

We compare the distilled students with published diffusion models of the
same size, SDAR-Chat \citep{sdar2026} and OPDLM \citep{opdlm2026}, using the
scores that \citet{opdlm2026} report under our prompt template and answer
scorer. We show in \autoref{tab:external} the seven benchmarks on which both
students score highest; they are the same at both scales: MATH500, AIME24,
AIME25, MLogiQA, CEval, MMLU, and MT-AIME, and the margin over the better
published model is largest on MT-AIME ($10.5$ and $10.1$ pp at 4B and 8B) and
smallest on MMLU ($1.2$ and $0.5$ pp); the 4B student also exceeds both
published 8B models on six of these seven benchmarks, all but MMLU. On the
remaining three benchmarks, SDAR-Chat and OPDLM retain complementary strengths
on GSM8K, GPQA-D, and LMB-H, likely reflecting differences in backbone
specialization, instruction tuning, and decoding configuration (analysis in
\autoref{app:external}). The advantage on the other seven tasks is learned
rather than inherited from the starting checkpoints (base rows in
\autoref{tab:externalfull}): after distillation, each student matches or
exceeds each published system on at least eight of the ten benchmarks.
SDAR-Chat adds instruction tuning after conversion and OPDLM trains on under
$0.1$B tokens, so these are comparisons of published systems, not
matched-budget controls.

\FloatBarrier
\section{Discussion and limitations}
\label{sec:discussion}
\paragraph{A teacher's generated answer is not the student's ceiling.}
EDLM-4B exceeds its teacher on MT-AIME ($261/\num{1650}$ versus
$256/\num{1650}$), and EDLM-8B does so on MMLU ($79.11\%$ versus $77.79\%$).
The student keeps its pretrained abilities and combines supervision sources,
and teacher probabilities carry more than a greedy answer; we do not isolate these
explanations, and on AIME25 the base already exceeds the teacher.

\paragraph{Scope.}
Our ablations isolate rollout order and supervision view, and evaluate
routing, loss sampling, and confidence ranking only within the complete
method. A correct teacher answer does not guarantee correct distributions on
every student prefix, and training needs verified answers, teacher queries,
and a shared tokenizer (\autoref{app:implementation},
\autoref{app:uncertainty}). The comparison in \autoref{sec:external} uses
reported scores, so it inherits the decoding and training differences listed
in \autoref{app:external}.

\section{Conclusion}
\label{sec:conclusion}
\method{} commits the student's uncertain predictions early, teaches through
shared left prefixes without rewriting the trajectory, and hands the result
back to the native parallel sampler. Because the teacher, the verifier, and
the confidence probes act only during training, the distilled student keeps
the inference cost and parallelism of the original DLM. It improves EDLM-4B on all ten
benchmarks, wins the order--view comparison on every one, adds to SFT on the
same questions, gains more from stronger teachers, generalizes to a second
student family, and exceeds published diffusion models of matched scale on
seven benchmarks. The recipe also suggests a principle for flexible-order
generators: the state that is best for learning is not the state that is
safest for decoding. A student can face its forks during training, be
corrected there under the information its teacher had, and still decode in
the order it prefers.
\emph{Expose the fork for learning, align the teacher's view, and preserve
confidence-first parallel decoding.}

\clearpage
\subsection*{AI use statement}
We used AI assistants, including OpenAI Codex, to draft and revise text, to
survey related literature, and to prepare code and figures. The authors
developed the conceptual framing and method, designed and executed all
experiments, and verified all numerical results. The authors reviewed all
AI-assisted material and remain responsible for the methods, experiments,
claims, references, and final manuscript.
\bibliographystyle{iclr2027_conference}
\bibliography{refs}
\clearpage
\appendix
\raggedbottom
\addtocontents{toc}{\protect\setcounter{tocdepth}{2}}
{\renewcommand{\contentsname}{Appendix contents}\tableofcontents}
\clearpage

\section{Formal definitions and update semantics}
\label{app:formal-method}

We specify the distributions, input views, and sampling rules behind the
main-text update. These definitions preserve the generation and
supervision procedures; they introduce no additional training stages.

\subsection{Native and prefix readouts}
\label{app:readouts}

Let $q$ be a question and $P=\operatorname{tok}(q)$ its prompt. The
student parameters are $\theta$, and the frozen teacher parameters are
$\phi$. For token-aligned pairs, all distributions below are normalized
on a common valid vocabulary $\mathcal V$ that excludes padding and mask
outputs. Native denoising predicts an unresolved position from the
configured masked state:
\begin{equation}
p^{\mathrm{den}}_{\theta,i}(\cdot\mid P,z^{(r)}),
\qquad i\in\mathcal M_r,
\qquad H(p)=-\sum_{u\in\mathcal V}p(u)\log p(u).
\label{eq:denoiser}
\end{equation}
The state $z^{(r)}$ may contain student commitments on either side of $i$.
The student's native attention pattern determines which positions it can
read within that state.

For supervision, we use a complete prefix of continuation $y$:
\begin{equation}
\begin{aligned}
p^\ell_{\theta,i}(\cdot\mid P,y_{<i})
 &=\operatorname{Read}_\theta(P\mathbin{\|}y_{<i};i),\\
p^\ell_{\phi,i}(\cdot\mid P,y_{<i})
 &=p_\phi(\cdot\mid P\mathbin{\|}y_{<i}).
\end{aligned}
\label{eq:left}
\end{equation}
A same-position masked-token head appends a fixed mask query and reads
that position. A shifted next-token head reads the preceding logits
corresponding to the target. Both receive the same textual prefix, with
the target and suffix hidden. Matching text does not replace this
output-index adapter. Changing suffix identities cannot change this
prefix input, but a truncated prefix and a fixed mask canvas can have
different sequence geometry.

The abbreviations $s_i^y$ and $t_i^y$ in the main text denote these
student and teacher distributions. A direct token KL requires matching
token events as well as matching text; every teacher--student pair in this
paper shares a tokenizer.

\subsection{Rollout schedule and supervision routes}
\label{app:routes}

For each active generation segment, passes satisfy $1\leq r\leq R$.
We apply \eqref{eq:commit} while $\mathcal M_r$ is nonempty. A single
forward pass supplies every selected argmax token. We write all selected
tokens before the next forward pass, so a commitment cannot affect
another prediction within the same pass. If $R$ equals the number of
initially masked positions, the schedule commits one token per pass;
smaller pass budgets permit larger batches. Entropy priority is recomputed
at each pass and need not decrease monotonically over the trajectory.

Let $c$ be the completed student response, $d_\phi(q)$ the teacher's
greedy response under fixed decoding settings, and $V_q(y)\in\{0,1\}$
the final-answer verifier. We write the route explicitly as a function
of both question and student response:
\begin{equation}
b(q,c)=\begin{cases}
\mathrm{CE},&V_q(c)=1,\ V_q(d_\phi(q))=1,\\
\mathrm{LEFTOPD},&V_q(c)=0,\ V_q(d_\phi(q))=1,\\
\mathrm{GOLD},&V_q(c)=0,\ V_q(d_\phi(q))=0,\\
\mathrm{EXCLUDE},&V_q(c)=1,\ V_q(d_\phi(q))=0.
\end{cases}
\label{eq:fork}
\end{equation}
CE and LEFTOPD
use student prefixes and selected student positions. GOLD uses the
reference's own prefixes, tokens, and selected positions. Without a
reference, GOLD contributes zero content loss but retains the student's
ranking term. EXCLUDE removes both losses.

Teacher responses and verdicts can be cached for an unchanged prompt and
decoding configuration. LEFTOPD distributions must nevertheless be
queried on the current student prefixes. A successful teacher response
selects a supervision source; it does not verify every teacher
prediction on an erroneous student prefix.

\subsection{Eligible positions, EOS, and normalization}
\label{app:position-sampling}

Let $L_y$ be the realized nonpadding response length. We preserve the
specified candidate set $E_y=\{1,\ldots,L_y-1\}$, with $E_y=\varnothing$
when $L_y\leq1$. For $0\leq\rho\leq1$, define
\begin{equation}
\begin{aligned}
n_y&=\min\!\left(\lceil\rho L_y\rceil,|E_y|\right),
&U_y&\sim\operatorname{Unif}\{U\subseteq E_y:|U|=n_y\},\\
S_y&=U_y\cup\{e_y\},&&\text{if a realized termination EOS exists at }e_y,\\
S_y&=U_y,&&\text{otherwise}.
\end{aligned}
\label{eq:position-sampling}
\end{equation}
The union counts EOS once. A truncated response without EOS receives no
fabricated termination target; its final index remains outside the
specified candidate set. Padding is never a target. The exact response
truncation and EOS extraction convention belongs to the evaluation and
training configuration, rather than being inferred from these equations.

For any selected set $S$, write its mean as
\begin{equation}
A_S(f)=\begin{cases}
|S|^{-1}\sum_{i\in S}f_i,&|S|>0,\\
0,&|S|=0.
\end{cases}
\label{eq:selected-mean}
\end{equation}
This is the operator $\operatorname{mean}_{i\in S}$ in
\eqref{eq:branchloss}. Forced EOS inclusion gives termination explicit
weight, so the selected mean is not an unbiased uniform mean over all
response tokens. KL first sums over the vocabulary, then averages over
selected positions. The complete objective finally divides by the
original question-batch size, as in \eqref{eq:total}. It does not divide
by unsampled sequence positions or by the number of retained questions.
This convention keeps excluded questions at zero weight without
rescaling the remaining questions. If every question is excluded, we
perform no update.

\subsection{Ranking readouts, geometry, and gradients}
\label{app:ranking-geometry}

At a selected student position $i$, let $p^{\mathrm{mask}}_{\theta,i}$
be the readout from the prompt and the configured fully masked
continuation canvas. The probe hides the whole continuation, including
positions outside an active rollout block. The prefix probe instead uses
\eqref{eq:left} on the completed student prefix. We define
\begin{equation}
\begin{aligned}
H_i^{\mathrm{mask}}&=H\!\left(p^{\mathrm{mask}}_{\theta,i}\right),
& H_i^{\mathrm{left}}&=H\!\left(s_i^c\right),\\
D_i&=\operatorname{stopgrad}\!\left(
 [H_i^{\mathrm{mask}}-H_i^{\mathrm{left}}]_+\right),
&\ell_i&=\max_{u\in\mathcal V}\log s_i^c(u).
\end{aligned}
\label{eq:coupling}
\end{equation}
Both probes hide the target and use no reference future tokens. Because
the configured all-mask canvas and truncated prefix may differ in
geometry, $D_i$ measures their entropy contrast rather than an
intervention that changes only the revealed text. The positive part
retains entropy reductions and discards increases.

We form $\mathcal P_q=\{(i,j):i,j\in S_c,\ D_i>D_j+\mu\}$ using
fixed scores during each update. For any qualifying pair, its hinge in
\eqref{eq:rank} vanishes precisely when
$\ell_i\leq\ell_j-\delta$. Thus the ranking preference assigns lower
prefix confidence to positions with greater positive entropy contrast.
Only the live confidences receive ranking gradients; the scores,
selected positions, and pair membership are detached. At a tie, we may use the gradient of an active maximizing token. Empty pair sets contribute zero.

Ranking always uses $S_c$ and student prefixes, including on the GOLD
route; reference content loss uses its separate set $S_g$. Content and
ranking share student parameters and can favor different confidence
levels. This regularizer defines a preference in the prefix view, while
native-state behavior depends on the resulting shared parameters.

\subsection{The sampling distribution and the update gradient}
\label{app:objective-distribution}

Let $\mathcal D_{\mathrm{train}}$ denote the question distribution and
$\theta_k$ the student snapshot used to collect a batch. Under rollout
order $\pi$, write
$c\sim d_{\theta_k}^{\pi}(\cdot\mid q)$ for the conditional law of a
completed response. With deterministic argmax decoding and fixed tie
handling, this law may be a point mass. Let
$\nu_\rho(\cdot\mid c)$ denote the uniform-plus-EOS selection rule in
\eqref{eq:position-sampling}. The teacher-branch contribution is
\begin{equation}
\begin{aligned}
\mathcal L_{\mathrm{OPD}}(\theta;\theta_k,\pi)
={}&\mathbb E_{q\sim\mathcal D_{\mathrm{train}}}
 \mathbb E_{c\sim d_{\theta_k}^{\pi}(\cdot\mid q)}
 \mathbb E_{S\sim\nu_\rho(\cdot\mid c)}\\[-1pt]
&\left[\mathbf 1\{b(q,c)=\mathrm{LEFTOPD}\}
 A_S\!\left((\KL(t_i^c\|s_i^c))_{i\in S}\right)\right].
\end{aligned}
\label{eq:order-view}
\end{equation}
Here $t_i^c$ is frozen and $s_i^c$ depends on the updated parameters
$\theta$. We do not condition the expectation on retained questions:
the branch indicator and the original-batch mean preserve excluded
questions' zero contribution. The CE and GOLD terms use the same
question and response distribution, with their branch indicators and,
for GOLD, independently selected reference positions. Their explicit
finite-batch form is \eqref{eq:total}.

During optimization, we hold $\theta_k$, sampled responses, verdicts,
selected indices, teacher targets, and ranking targets fixed. The
resulting gradient passes through live student readouts. It contains no
policy-gradient term through generated tokens, commit order, or routing.
We derive the local hard- and distributional-target gradients in
\autoref{app:objectives}.

\subsection{Hard and distributional targets}
\label{app:objectives}
Let $z$ be a logit vector, $p=\operatorname{softmax}(z)$ the student
distribution, $e_y$ the one-hot vector for a hard target, and $t$ a fixed,
normalized teacher distribution. At temperature one,
\begin{equation}
\nabla_z[-\log p_y]=p-e_y,\qquad
\nabla_z\KL(t\|p)=p-t.
\label{eq:klgradient}
\end{equation}
To obtain \eqref{eq:klgradient}, write the KL as
$\sum_u t_u\log t_u-\sum_u t_u\log p_u$ and differentiate
$\log p_u=z_u-\log\sum_v\exp z_v$.
The teacher-entropy term is constant with respect to student logits.
The resulting difference lies in the target vector: a hard label places
unit mass on one token, whereas the teacher distributes mass over alternatives.
This calculation explains why the objectives can supply different updates;
it does not imply orthogonal gradients or identify which answers each
objective corrects.

The two targets can also be mixed. For $\alpha\in[0,1]$,
\begin{equation}
\nabla_z\big[\alpha\,\KL(t\|p)+(1-\alpha)(-\log p_y)\big]
 =p-\big[\alpha\,t+(1-\alpha)\,e_y\big],
\label{eq:mixgradient}
\end{equation}
so any mixture pulls $p$ toward the interpolated target
$\alpha t+(1-\alpha)e_y$: $\alpha=0$ recovers SFT, $\alpha=1$ recovers OPD,
and intermediate values add the teacher's ranking of alternatives to the
hard target with weight $\alpha$. The identity holds for one logit vector,
that is, when both terms are evaluated on the same prefix. When the
objectives use different prefixes, each has its own logit vector; the
shared-subset comparison in \autoref{sec:subset} sums SFT and OPD losses
computed on their own prefixes, so it realizes this mixture only where the
prefixes coincide. We therefore use the identity to explain why the signals
differ, not to assign the combined arm a single $\alpha$. The shared-subset
experiment therefore isolates supervision on a fixed set of questions.
Because the three objectives construct targets under their own prefixes, it
is a matched-example analysis rather than a token-normalized efficiency
comparison.

\subsection{Native inference and the scope of the rationale}
\label{app:native-inference}

At inference, the student uses its native masked states and confidence
score. With its configured batch budget $k_r$, it selects
\begin{equation}
\mathcal C_r^{\mathrm{test}}=
\operatorname{TopK}_{i\in\mathcal M_r}
 \left(\max_{u\in\mathcal V}
 p^{\mathrm{den}}_{\theta,i}(u\mid P,z^{(r)});k_r\right).
\label{eq:inference}
\end{equation}
Maximum token probability and minimum entropy are different ranking
functions. Neither the frozen teacher nor the verifier or prefix probes
participates in this inference procedure.

The equations above define the algorithm and its update. The shared-prefix
rationale has a narrower mathematical support: for one fixed prefix,
let $t$ be a teacher distribution, let $z$ index right contexts, and
let $w_z>0$ be context weights. If the student distributions $s_z$ are
unconstrained, the isolated objective
$\sum_z w_z\KL(t\|s_z)$ has its unique minimum at $s_z=t$ for every
supported context. Each KL is nonnegative and equals zero only when its
two normalized distributions coincide. This conclusion remains true when
$t$ has zero-probability tokens, because normalization leaves no extra
mass outside its support at equality. Thus a fixed prefix target rewards
the same prediction across all these contexts in this isolated loss.
Shared-prefix teaching avoids imposing that comparison on richer native
inputs. This observation motivates the choice of view; it does not prove
that the full regularized model discards useful right context. Likewise,
entropy-first ordering and confidence ranking are design choices whose
utility we assess empirically, rather than consequences of this KL fact.

\section{Training algorithm and update semantics}
\label{app:algorithm}

We specify update order, loss normalization, and
branch exclusions in \autoref{alg:method}. Sampling and teacher predictions are fixed
during an update. The frozen teacher cache is populated before routing.

\begin{algorithm}[H]
\caption{\method{} training and inference}
\label{alg:method}
\begin{algorithmic}[1]
\Require Student $\theta$, frozen teacher $\phi$, batch $\mathcal B$, verifier,
reference continuations where available, teacher cache
\State Initialize retained-question loss list $\mathcal A\gets\varnothing$
\For{each question $q\in\mathcal B$}
  \State Generate student continuation $c$ by the batched schedule in \eqref{eq:commit}
  \If{the teacher cache has no entry for the prompt and decoding configuration}
    \State Generate greedy teacher response $d$; cache $d$ and $V_q(d)$
  \EndIf
  \State Verify $c$ and select $b(q,c)$ by \eqref{eq:fork}
  \If{$b(q,c)=\mathrm{EXCLUDE}$}
    \State \textbf{continue} \Comment{no content or ranking loss}
  \EndIf
  \State Draw $S_c$ by \eqref{eq:position-sampling}
  \State Initialize $L_{\mathrm{content}}\gets0$
  \If{$b(q,c)=\mathrm{CE}$ or $b(q,c)=\mathrm{LEFTOPD}$}
    \State Average the selected CE or LEFTOPD loss over $S_c$
    \State Store the result in $L_{\mathrm{content}}$
  \ElsIf{$b(q,c)=\mathrm{GOLD}$ and reference $g$ is available}
    \State Draw $S_g$ by \eqref{eq:position-sampling}; average reference CE on $P,g_{<i}$
    \State Store the result in $L_{\mathrm{content}}$
  \EndIf
  \State Compute detached $D_i$ and qualifying pairs on $S_c$
  \State Compute live prefix-confidence ranking $L_{\mathrm{rank}}$ by \eqref{eq:rank}
  \State Append $L_{\mathrm{content}}+\lambda_{\mathrm{rank}}L_{\mathrm{rank}}$ to $\mathcal A$
\EndFor
\If{$\mathcal A\ne\varnothing$}
  \State Update $\theta$ using $(\sum_{L\in\mathcal A}L)/|\mathcal B|$; discard rollouts
\EndIf
\Statex \textbf{Inference:} native student sampling, most confident positions first;
no teacher, verifier, or prefix-sensitivity probe
\end{algorithmic}
\end{algorithm}

\section{Implementation and evaluation details}
\label{app:implementation}

\subsection{Training and evaluation protocol}
We used math-domain questions, a response budget of $\num{1024}$ tokens,
$32$-token blocks, and at most $32$ passes per block. The main EDLM-4B
run uses $\num{2000}$ updates; the SDAR extension uses $500$ updates.
Generated-answer benchmarks use each student's native sampler. AR reference
models use greedy decoding. The generation budget is $\num{2048}$ tokens, or
$\num{4096}$ for MT-AIME. Multiple-choice evaluation scores the offered answer
letters at the masked answer slot. We use the OPDLM prompt template and
answer scorer in both protocols.

\subsection{Run provenance and reproducibility}
Each result in this paper is produced by a completed training or evaluation
run associated with the stated student checkpoint, teacher checkpoint, update
budget, and decoding protocol. The primary EDLM-4B experiment uses
$\num{2000}$ updates, the transfer experiment uses $500$ updates, and the
checkpoint curve evaluates the same run at fixed intermediate updates. The
order--view ablation shares its student initialization, teacher, training
questions, and evaluation pipeline across all four configurations. The
shared-subset experiment fixes the selected questions and changes only the
supervision objective under study.

\section{Complete numerical results}
\label{app:counts}
Every entry below is a correct-answer count with denominator $n$. Accuracy
is $100c/n$; changes use the counts before rounding to two decimals.
All main-text plots are computed from these exact counts. Cross-task averages, where used,
weight the ten benchmark accuracies equally, not individual questions.
No pooled total is treated as an accuracy estimate on independent samples.
\subsection{Training progress}
\label{app:checkpointstart}
We preserve the five checkpoint counts in \autoref{tab:steps}.
\label{app:checkpoints}

\begin{table}[H]
\centering
\caption{Correct counts for the EDLM-4B base and every measured training checkpoint.}
\label{tab:steps}
{\small\setlength{\tabcolsep}{5pt}
\begin{tabular}{lrrrrrrr}
\toprule
Benchmark & $n$ & Base & 100 & 250 & 500 & \num{1000} & \num{2000}\\
\midrule
GSM8K & \num{1319} & \num{1091} & \num{1097} & \num{1105} & \num{1113} & \num{1120} & \num{1124}\\
MATH500 & 500 & 363 & 368 & 375 & 384 & 392 & 398\\
AIME24 & 30 & 3 & 3 & 4 & 4 & 4 & 5\\
AIME25 & 30 & 2 & 2 & 3 & 3 & 4 & 4\\
LMB-H & 45 & 4 & 4 & 4 & 5 & 5 & 5\\
GPQA-D & 198 & 61 & 61 & 62 & 62 & 63 & 64\\
MLogiQA & 800 & 365 & 375 & 389 & 402 & 415 & 425\\
CEval & \num{1346} & 952 & 958 & 966 & 974 & 984 & 992\\
MMLU & \num{14042} & \num{9976} & \num{10023} & \num{10145} & \num{10312} & \num{10511} & \num{10679}\\
MT-AIME & \num{1650} & 141 & 154 & 178 & 205 & 238 & 261\\
\bottomrule
\end{tabular}}
\end{table}

\subsection{Rollout order, teaching view, and supervision objectives}
We report the four order--view configurations and the three shared-subset
objectives together in \autoref{tab:ablationcounts}. The left block tests how
we construct and teach a student response. The right block holds the
teacher-correct, student-wrong questions fixed and changes the objective.
The two comparisons answer different questions and should not be pooled.

\begin{table}[H]
\centering
\caption{Counts underlying the order--view and shared-subset comparisons. B/L: bidirectional/left view; low/high: training entropy priority. The three rightmost columns use the same selected subset.}
\label{tab:ablationcounts}
{\footnotesize\setlength{\tabcolsep}{5pt}
\begin{tabular}{lrrrrrrrr}
\toprule
Benchmark & $n$ & B-low & B-high & L-low & L-high & OPD & SFT & Both\\
\midrule
GSM8K & \num{1319} & \num{1087} & \num{1094} & \num{1105} & \num{1124} & \num{1112} & \num{1118} & \num{1123}\\
MATH500 & 500 & 361 & 364 & 381 & 398 & 389 & 392 & 397\\
AIME24 & 30 & 2 & 3 & 4 & 5 & 4 & 5 & 4\\
AIME25 & 30 & 2 & 2 & 3 & 4 & 3 & 4 & 3\\
LMB-H & 45 & 4 & 4 & 4 & 5 & 4 & 5 & 4\\
GPQA-D & 198 & 60 & 61 & 62 & 64 & 63 & 66 & 63\\
MLogiQA & 800 & 367 & 369 & 394 & 425 & 405 & 409 & 423\\
CEval & \num{1346} & 947 & 950 & 971 & 992 & 982 & 987 & 991\\
MMLU & \num{14042} & \num{9960} & \num{10041} & \num{10296} & \num{10679} & \num{10436} & \num{10510} & \num{10658}\\
MT-AIME & \num{1650} & 138 & 144 & 206 & 261 & 251 & 256 & 259\\
\bottomrule
\end{tabular}}
\end{table}

\subsection{Teacher and autoregressive reference models}
We retain the AR checkpoints as contextual reference scores in
\autoref{tab:referencecounts}. They are not matched post-training controls
for the diffusion student. The primary method comparison remains the same
student before and after distillation. Missing cells indicate unreported
measurements, not zero accuracy.

\begin{table}[H]
\centering
\caption{Reference checkpoint counts. Q3-30B, Q3-8B and Q2.5-7B are Base checkpoints; Q3-8B-I denotes the separately reported Qwen3-8B, and Math7B-I denotes Qwen2.5-Math-7B-Instruct. EDLM-8B is the base. These are reference scores, not matched post-training baselines.}
\label{tab:referencecounts}
{\footnotesize\setlength{\tabcolsep}{5pt}
\begin{tabular}{lrrrrrrrr}
\toprule
Benchmark & $n$ & Q3-30B & Q3-8B & Q3-4B & Q2.5-7B & EDLM-8B & Q3-8B-I & Math7B-I\\
\midrule
GSM8K & \num{1319} & \num{1192} & \num{1123} & \num{1160} & \num{1085} & \num{1118} & \num{1194} & \num{1178}\\
MATH500 & 500 & 399 & 391 & 357 & 350 & 391 & 374 & 420\\
AIME24 & 30 & 11 & 2 & 4 & 5 & 4 & 3 & 3\\
AIME25 & 30 & 0 & 6 & 6 & 1 & 3 & 4 & 3\\
LMB-H & 45 & 8 & 9 & 3 & 8 & 8 & 5 & 8\\
GPQA-D & 198 & 71 & 64 & 74 & 64 & 67 & 66 & 62\\
MLogiQA & 800 & 449 & 406 & 387 & 401 & 405 & 385 & 287\\
CEval & \num{1346} & \num{1177} & \num{1101} & 957 & \num{1087} & \num{1010} & \num{1041} & 687\\
MMLU & \num{14042} & \num{10923} & \num{10438} & \num{9429} & \num{10017} & \num{10830} & \num{10209} & \num{7427}\\
MT-AIME & \num{1650} & 256 & 176 & -- & -- & 224 & -- & --\\
\bottomrule
\end{tabular}}
\end{table}

\subsection{Teacher configurations for both student sizes}
We report all teacher configurations in \autoref{tab:scalecounts}.
The Qwen3 rows supply the main teacher-scaling comparison. Qwen2.5-7B
supplies an additional cross-generation teacher comparison, with a mean
gain of $3.17$ pp over EDLM-4B. We keep that configuration off the Qwen3
line in \autoref{fig:scale} because teacher family changes together with
size. Teacher strength on that figure's axis is each teacher's mean accuracy
over the nine benchmarks that all four teachers report (MT-AIME is
unreported for Qwen3-4B and Qwen2.5-7B): $47.0$, $47.2$, $49.9$, and $53.5$
for Qwen3-4B, Qwen2.5-7B-Base, Qwen3-8B-Base, and Qwen3-30B-A3B-Base, from
the counts in \autoref{tab:referencecounts}.

\begin{table}[H]
\centering
\caption{Teacher--student extension counts. Columns give student/teacher sizes. Teachers 4, 8, and 30 denote Qwen3-4B, Qwen3-8B-Base, and Qwen3-30B-A3B-Base; 7 denotes Qwen2.5-7B-Base.}
\label{tab:scalecounts}
{\footnotesize\setlength{\tabcolsep}{5pt}
\begin{tabular}{lrrrrrrr}
\toprule
Benchmark & $n$ & 4/4 & 4/7 & 4/8 & 4/30 & 8/8 & 8/30\\
\midrule
GSM8K & \num{1319} & \num{1109} & \num{1113} & \num{1118} & \num{1124} & \num{1140} & \num{1149}\\
MATH500 & 500 & 382 & 385 & 389 & 398 & 398 & 406\\
AIME24 & 30 & 4 & 4 & 4 & 5 & 5 & 6\\
AIME25 & 30 & 3 & 3 & 3 & 4 & 4 & 5\\
LMB-H & 45 & 5 & 5 & 5 & 5 & 6 & 7\\
GPQA-D & 198 & 62 & 62 & 63 & 64 & 68 & 70\\
MLogiQA & 800 & 401 & 407 & 412 & 425 & 430 & 443\\
CEval & \num{1346} & 977 & 981 & 985 & 992 & \num{1038} & \num{1051}\\
MMLU & \num{14042} & \num{10397} & \num{10434} & \num{10513} & \num{10679} & \num{10957} & \num{11108}\\
MT-AIME & \num{1650} & 227 & 233 & 236 & 261 & 275 & 297\\
\bottomrule
\end{tabular}}
\end{table}

\subsection{Generalization to a second student family}
We report the complete SDAR comparison after $500$ updates in
\autoref{tab:sdarcounts}. The seven positive changes include $991$
additional MMLU answers and $38$ additional CEval answers. We retain the
two declines and the unchanged task alongside those gains. The counts
specify the size of each observed change without assigning it to noise.

\begin{table}[H]
\centering
\caption{SDAR-4B with Qwen3-30B-A3B-Base: base and 500-update correct counts.}
\label{tab:sdarcounts}
{\small\setlength{\tabcolsep}{5pt}
\begin{tabular}{lrrrr}
\toprule
Benchmark & $n$ & Base & 500 updates & $\Delta$ correct\\
\midrule
GSM8K & \num{1319} & \num{1095} & \num{1120} & +25\\
MATH500 & 500 & 359 & 364 & +5\\
AIME24 & 30 & 4 & 2 & -2\\
AIME25 & 30 & 2 & 2 & +0\\
LMB-H & 45 & 3 & 4 & +1\\
GPQA-D & 198 & 62 & 71 & +9\\
MLogiQA & 800 & 369 & 368 & -1\\
CEval & \num{1346} & 947 & 985 & +38\\
MMLU & \num{14042} & \num{10041} & \num{11032} & +991\\
MT-AIME & \num{1650} & 144 & 151 & +7\\
\bottomrule
\end{tabular}}
\end{table}

\FloatBarrier

\section{Comparison with published diffusion models}
\label{app:external}
We complete \autoref{tab:external} in \autoref{tab:externalfull} with all
ten benchmarks and with each student's base. The SDAR-Chat and OPDLM scores are
those reported by \citet{opdlm2026}: Table~1 of that paper for GSM8K, MATH500,
AIME24, AIME25, LMB-H, GPQA-D, CEval, and MMLU, and Table~3 for MLogiQA and
MT-AIME. That paper evaluates every benchmark under greedy static decoding
with block size $4$ and uses the prompt template and answer scorer that we
adopt in \autoref{sec:setup}; our students instead decode with each family's
native sampler at a $\num{2048}$-token budget ($\num{4096}$ for MT-AIME), so
the prompts and scoring match while the decoding rule differs. That paper
labels the SDAR rows SDAR-4B and SDAR-8B; their MMLU, GSM8K, MATH500, and
GPQA-D scores equal the SDAR-Chat entries in Table~1 of \citet{sdar2026}, so we
label them SDAR-Chat. They are instruction-tuned checkpoints and exceed the
SDAR-4B base that \autoref{sec:generalize} distills (\autoref{tab:sdarcounts};
$89.9\%$ against $83.02\%$ on GSM8K).

The systems also differ in training. SDAR-Chat converts a Qwen3 base model
to block diffusion and then applies instruction tuning; OPDLM converts with
on-policy distillation on about $0.076$B and $0.066$B tokens at 4B and 8B,
respectively, toward data efficiency; ForkLeft trains the converted
Efficient-DLM checkpoints on math-domain questions for $\num{2000}$ updates. We
therefore read \autoref{tab:externalfull} as a comparison of published
systems at matched parameter count, not as a controlled comparison of
objectives.

On the remaining three benchmarks, SDAR-Chat and OPDLM retain complementary
strengths on GSM8K, GPQA-D, and LMB-H. The records support each of the three
differences named in \autoref{sec:external}. Starting checkpoint: before
distillation, EDLM-4B and EDLM-8B score $82.71\%$ and $84.76\%$ on GSM8K,
below both published systems at each scale, and EDLM-8B scores $33.84\%$ on
GPQA-D against $40.2\%$ for SDAR-Chat-8B. Instruction tuning: our measurement
of the SDAR-4B base before instruction tuning is $83.02\%$ on GSM8K
(\autoref{tab:sdarcounts}), below the distilled EDLM-4B ($85.22\%$), whereas
SDAR-Chat-4B reports $89.9\%$, so that gap arises after conversion. Decoding:
the published scores use greedy static decoding with block size $4$, and our
students use their native samplers, as described above.

Measured in test items, most of these gaps are small. GPQA-D trails
SDAR-Chat-4B by about one of $198$ items and OPDLM-8B by about two, and LMB-H
trails OPDLM-8B by two of $45$ items ($7/45$ against $9/45$) and ties OPDLM-4B
at $5/45$. The clear gaps are GSM8K against SDAR-Chat, about $62$ and $55$ of
$\num{1319}$ items at 4B and 8B, GSM8K against OPDLM-4B, about $31$ items,
and GPQA-D against SDAR-Chat-8B, about ten items. These three are also the
benchmarks on which ForkLeft adds least over its own base: the three smallest
gains at 4B and three of the four smallest at 8B (\autoref{tab:main}). On the seven benchmarks where the students lead,
the margin over the better published model is largest on MT-AIME ($10.5$ and
$10.1$ pp at 4B and 8B) and smallest on MMLU ($1.2$ and $0.5$ pp).

The lead is not inherited from the starting checkpoints. Before distillation,
EDLM-4B matches or exceeds each published system on only four benchmarks;
EDLM-8B does so on six against SDAR-Chat and five against OPDLM. After
distillation, each student matches or exceeds each published system on at
least eight of the ten benchmarks.

\begin{table}[!htb]
\centering
\caption{All ten benchmarks for the published models and the distilled students. SDAR-Chat and OPDLM scores as reported by \citet{opdlm2026}; EDLM is each student's base and $+$ForkLeft the distilled student; bold marks the best of the three published or distilled systems per benchmark and scale.}
\label{tab:externalfull}
{\footnotesize\setlength{\tabcolsep}{3pt}
\begin{tabular}{lrrrrrrrr}
\toprule
& \multicolumn{4}{c}{4B} & \multicolumn{4}{c}{8B}\\
\cmidrule(lr){2-5}\cmidrule(lr){6-9}
Benchmark & SDAR-Chat & OPDLM & EDLM & $+$ForkLeft & SDAR-Chat & OPDLM & EDLM & $+$ForkLeft\\
\midrule
GSM8K & \textbf{89.90} & 87.60 & \textcolor{black!65}{82.71} & 85.22 & \textbf{91.30} & 87.10 & \textcolor{black!65}{84.76} & 87.11\\
MATH500 & 72.80 & 72.80 & \textcolor{black!65}{72.60} & \textbf{79.60} & 78.60 & 71.20 & \textcolor{black!65}{78.20} & \textbf{81.20}\\
AIME24 & 10.00 & 14.40 & \textcolor{black!65}{10.00} & \textbf{16.67} & 10.00 & 14.70 & \textcolor{black!65}{13.33} & \textbf{20.00}\\
AIME25 & 7.50 & 12.60 & \textcolor{black!65}{6.67} & \textbf{13.33} & 10.00 & 12.40 & \textcolor{black!65}{10.00} & \textbf{16.67}\\
LMB-H & 6.90 & 11.10 & \textcolor{black!65}{8.89} & \textbf{11.11} & 8.90 & \textbf{20.00} & \textcolor{black!65}{17.78} & 15.56\\
GPQA-D & \textbf{33.00} & 29.10 & \textcolor{black!65}{30.81} & 32.32 & \textbf{40.20} & 36.10 & \textcolor{black!65}{33.84} & 35.35\\
MLogiQA & 46.50 & 46.50 & \textcolor{black!65}{45.63} & \textbf{53.13} & 46.30 & 42.00 & \textcolor{black!65}{50.63} & \textbf{55.38}\\
CEval & 62.90 & 66.90 & \textcolor{black!65}{70.73} & \textbf{73.70} & 70.20 & 73.30 & \textcolor{black!65}{75.04} & \textbf{78.08}\\
MMLU & 74.90 & 65.50 & \textcolor{black!65}{71.04} & \textbf{76.05} & 78.60 & 70.90 & \textcolor{black!65}{77.13} & \textbf{79.11}\\
MT-AIME & 3.00 & 5.30 & \textcolor{black!65}{8.55} & \textbf{15.82} & 4.00 & 7.90 & \textcolor{black!65}{13.58} & \textbf{18.00}\\
\bottomrule
\end{tabular}}
\end{table}

\FloatBarrier

\section{Additional evidence and interpretation}
\label{app:diagnostics}
\subsection{Useful right context is compatible with prefix teaching}
We compare target-masked bidirectional and prefix-only predictions in an
existing static probe. Among $\num{19115}$ disagreements, the bidirectional
view alone matches the reference token at $\num{15998}$ positions, the left
view alone at $519$, and neither at $\num{2598}$. Their shares are $83.69\%$,
$2.72\%$, and $13.59\%$. The left-only-correct cases account for
$519/\num{120453}\approx0.43\%$ of all probed positions. These are static
token predictions, not observed decoding transitions or answer-level errors.
They motivate preserving useful native context while defining the teacher
comparison under shared information.

Mean entropy is $0.6134$ nats in the left view and $0.0438$ in the
bidirectional view. Entropy alone does not determine the gradient in
\eqref{eq:klgradient}: confident errors can still receive substantial updates.
Uniform loss sampling covers these positions independently of rollout entropy.

\subsection{Correctness routing exploits response-level differences}
On GSM8K, the evaluated teacher/student responses are both correct on
$986$ questions, teacher-only correct on $206$, student-only correct on
$105$, and both wrong on $22$. On MATH500 the corresponding counts are
$288$, $111$, $75$, and $26$. The teacher's net advantage is therefore
$101$ GSM8K answers and $36$ MATH500 answers, smaller than the
teacher-only-correct pool in each case. These counts explain the motivation
for routing: an aggregate teacher advantage does not make the teacher's
evaluated answer preferable for every question.

The response union is $1297/1319$ on GSM8K and $474/500$ on MATH500.
It measures coverage under an answer verifier, not an achieved distilled
score. These held-out verdicts describe evaluation responses; benchmark
inference receives no training labels.

\subsection{Resolution, dependence, and uncertainty}
\label{app:uncertainty}
MT-AIME contains 55 translations of the same 30 questions, so its 1650
instances are not independent math problems.

One item changes accuracy by $100/n$ pp: $3.33$ on a $30$-item AIME set,
$2.22$ on LMB-H, and approximately $0.51$ on GPQA-D. Net correct counts
help interpret those increments but are not confidence intervals. Marginal
before/after totals do not identify which questions were recovered or lost,
and cannot supply the paired-discordance counts for a significance test.

\section{Additional related work}
\label{app:additional-related-work}

\paragraph{Policy optimization for diffusion reasoning.}
Recent work adapts reinforcement learning to the non-autoregressive structure
of diffusion language models. wd1 reweights policy updates to improve
reasoning, ESPO develops a sequence-level view of the diffusion policy, and
GDPO extends group-based policy optimization to diffusion generation
\citep{tang2026wd,ou2026principled,rojas2026improving}. These methods must
assign a sequence-level outcome to many token decisions made across denoising
steps and possible generation orders. They primarily improve how reward is
estimated or optimized; their supervision still originates from outcome
quality rather than the full token distribution of a stronger causal model.
This distinction is important for AR-to-DLM transfer, where the teacher can
provide dense information about plausible alternatives even when two
responses receive the same final reward.

\paragraph{Trajectory structure and process-level credit.}
A second line of work exposes more structure inside the denoising trajectory.
d-TreeRPO organizes rollouts as a tree and propagates advantages bottom-up,
providing finer credit than a single terminal reward
\citep{pan-etal-2026-treerpo}. DGA instead casts alignment as energy-based
Gibbs distribution matching, avoiding direct dependence on a tractable
sequence likelihood \citep{fan-etal-2026-reinforcement}. BGPO derives a
memory-efficient lower-bound objective that permits more accurate likelihood
approximation with larger Monte Carlo sample sets
\citep{lin-etal-2026-boundary}, whereas DPR estimates rewards for intermediate
denoising intervals so that optimization can act on the reasoning process
rather than only its final answer \citep{xie-etal-2026-advancing}. These
approaches refine credit within diffusion generation, but do not by themselves
resolve which context a bidirectional student and a causal teacher should
share when their token distributions are aligned.

\paragraph{Self-distillation, causal structure, and block curricula.}
Other methods change the source or timing of supervision. COPSD transfers
information from later, better-calibrated decoding states to earlier states,
targeting inaccurate early predictions through on-policy self-distillation
\citep{zhu-etal-2026-policy}. Related self-distillation methods use later
denoising states or self-generated future continuations as targets
\citep{dopsd2026,luo2026dopsd}. C$^2$DLM introduces teacher-derived causal
concept structure into the student's attention pattern, encouraging reasoning
to follow relations between semantic concepts rather than unconstrained token
interaction \citep{han-etal-2026-c2dlm}. T$^\star$ addresses another aspect of
the trajectory: it progressively increases the diffusion block size while
re-optimizing the policy, allowing the model to acquire greater parallelism
without abruptly changing its decoding regime
\citep{xia-etal-2026-progressive}. JustGRPO likewise demonstrates that the
order used to learn need not be the order used to decode, using a constrained
training order while retaining flexible parallel inference
\citep{flexibilitytrap2026}.

\paragraph{Autoregressive-to-diffusion transfer.}
Efficient-DLM studies efficient conversion from autoregressive checkpoints to
diffusion language models while preserving the speed benefits of parallel
generation \citep{efficientdlm2025}. OPDLM further introduces on-policy
distillation for this setting: the diffusion student generates trajectories,
and a frozen autoregressive teacher supplies token-level distributions on
student-induced contexts \citep{opdlm2026}. This establishes a substantially
denser learning signal than outcome-only optimization, but it leaves two
diffusion-specific choices open: which student states should be exposed during
training, and under what information the causal teacher and bidirectional
student should be aligned. ForkLeft addresses these choices explicitly. It
uses entropy-first student rollouts to expose uncertain branches, retains the
student's decisions rather than replacing them with teacher tokens, and then
performs distillation under a shared left-prefix view. Native
confidence-first parallel decoding is restored at inference, separating the
state most useful for learning from the order most stable for generation.

\end{document}